%% file: main.tex
\PassOptionsToPackage{table}{xcolor}
\documentclass[10pt,twocolumn,letterpaper]{article}

\usepackage[pagenumbers]{cvpr} 


\definecolor{cvprblue}{rgb}{0.21,0.49,0.74}
\usepackage[pagebackref,breaklinks,colorlinks,allcolors=cvprblue]{hyperref}

\usepackage{pifont}
\usepackage{booktabs}
\usepackage{graphicx}
\usepackage{multirow}
\usepackage{makecell}
\usepackage{array}
\usepackage{tabularx}
\usepackage{rotating}
\usepackage{algorithm}
\usepackage{algpseudocode}
\usepackage{lipsum}
\usepackage[accsupp]{axessibility} 
\usepackage{amsmath}
\numberwithin{equation}{section}

\newcommand\blfootnote[1]{%
  \begingroup
  \renewcommand\thefootnote{}\footnote{#1}%
  \addtocounter{footnote}{-1}%
  \endgroup
}
\def\paperID{12752} 
\def\confName{CVPR}
\def\confYear{2025}

\title{V-Stylist: Video Stylization via Collaboration and Reflection of MLLM Agents}

\author{
    Zhengrong Yue$^{1,2}$,
    Shaobin Zhuang$^{1,2}$, 
    Kunchang Li$^{2,3,4}$,
    Yanbo Ding$^{3,4}$,
    Yali Wang$^{3,2\dag}$ \\
    \footnotesize{$^1$Shanghai Jiao Tong University
    \quad $^2$Shanghai AI Laboratory
    \quad} \\
    \footnotesize{$^3$Shenzhen Institutes of Advanced Technology, Chinese Academy of Sciences \quad} \\
    \footnotesize{$^4$University of Chinese Academy of Sciences \quad} 
}

\begin{document}
\maketitle
\input{sec/0_abstract}
\input{sec/1_intro}

\input{sec/2_related_work}
\input{sec/3_method}

\input{sec/4_experiment}
\input{sec/5_conclusion}
\input{sec/6_acknowledgments}

{
    \small
    \bibliographystyle{ieeenat_fullname}
    \bibliography{main}
}

\clearpage
\input{sec/X_suppl}



\end{document}

%% file: sec/0_abstract.tex
\begin{abstract}
Despite the recent advancement in video stylization, 
most existing methods struggle to render any video with complex transitions,
based on an open style description of user query.
To fill this gap,
we introduce a generic multi-agent system for video stylization, 
\textbf{V-Stylist}, 
by a novel collaboration and reflection paradigm of multi-modal large language models. 
Specifically, 
our V-Stylist is a systematical workflow with three key roles: 
(1) \textbf{Video Parser} decomposes the input video into a number of shots and generates their text prompts of key shot content.
Via a concise video-to-shot prompting paradigm,
it allows our V-Stylist to effectively handle videos with complex transitions. 
(2) \textbf{Style Parser} identifies the style in the user query and progressively search the matched style model from a style tree.
Via a robust tree-of-thought searching paradigm,
it allows our V-Stylist to precisely specify vague style preference in the open user query.
(3) \textbf{Style Artist} leverages the matched model to render all the video shots into the required style.
Via a novel multi-round self-reflection paradigm,
it allows our V-Stylist to adaptively adjust detail control,
according to the style requirement.
With such a distinct design of mimicking human professionals, 
our V-Stylist achieves a major breakthrough over the primary challenges for effective and automatic video stylization. 
Moreover,  
we further construct a new benchmark Text-driven Video Stylization Benchmark (TVSBench),
which fills the gap to assess various stylization of complex videos on open user queries. 
Extensive experiments show that, 
V-Stylist achieves the state-of-the-art,
e.g.,V-Stylist surpasses FRESCO and ControlVideo by 6.05\% and 4.51\% respectively in overall average metrics, 
marking a significant advance in video stylization.
\end{abstract}

%% file: sec/1_intro.tex
\section{Introduction}
\label{sec:intro}


\begin{figure}[tp]
    \centering
    \includegraphics[width=\linewidth
    ]{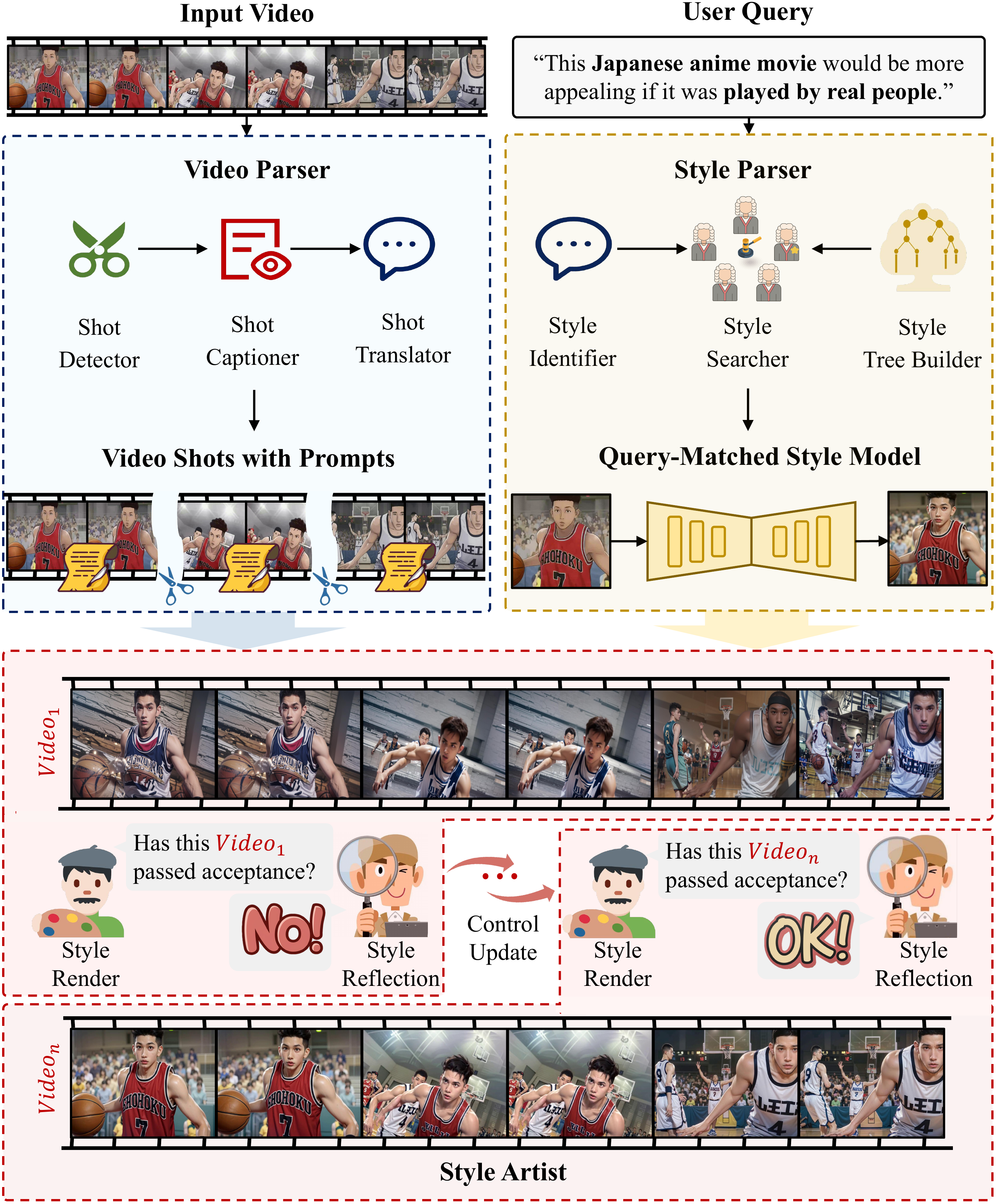}
    \vspace{-0.6cm}
    \caption{
    \textbf{Overview.} 
    V-Stylist is a multi-agent system for complex video stylization with open user query. 
    It contains three key roles to address the primary challenges in video stylization.
    Video Parser decomposes a video into several shots and prompts,
    via a video-to-shot prompting paradigm.
    Style Parser finds the most suitable stylization model,
    via a tree-of-thought searching paradigm.
    Style Artist adaptively adjusts detail control for preferable shot rendering,
    via a multi-round style reflection paradigm.
    }
    \label{fig:teaser}
    \vspace{-0.5cm}
\end{figure}

\blfootnote{$\dag$ Corresponding author.}

Video stylization is a classic and important topic in computer vision,
which has widespread applications in the real world, 
such as art design, education training, advertising, etc.
In the recent years,
with the fast development of text-to-image generation \cite{ddpm, ddim, stable-diffusion, imagen, dalle2},
the advancement of video stylization has been mainly driven by diffusion models \cite{gen1, vid2vid-zero, video-p2p, tokenflow, pix2video, fatezero, lovecon, controlvideo, control-a-video, fairy, flowvid, flatten},
for making a video with the required style of user query.

However,
the current approaches encounter three principal challenges in video stylization.
(1) \textbf{Complex Video Transitions.}
Most existing methods are designed for processing short video clips with only a few seconds. 
However,
the videos in our real life often contain complex scene transitions with significant motions in a longer period,
e.g., the original animated basketball video consists of three distinct scenes in \cref{fig:teaser}.
In such a case,
these methods are often infeasible or deteriorate significantly for stylization.
(2) \textbf{Vague Style Preference.}
Typically,
most existing methods directly use the user query as text prompt for style control.
However,
we notice that the general users are often not the professional artists.
Their queries are often ambiguous descriptions on style preference,  
e.g., the user query is \textit{``This Japanese anime movie would be more appealing if it was performed by real people"} in \cref{fig:teaser}, 
which contains both anime movie and real people but without explicit expression of exact style tag.
Such open prompts would mislead the models for poor stylization.  
(3) \textbf{Fixed Detail Control.}
Different styles necessitate different levels of control over spatial structure detail. 
For instance, 
the animation style favors strict line structure control, 
while 
the oil painting style prefers less rigid structure control for greater creative freedom.
However,
most current approaches ignore such an adaptive property,
and blindly control details with the fixed weight setting on a set of ControlNets.
This would lead to low-level visual distortion or high-level character shifting in the stylized videos.
Hence,
there is a natural question: 
\textit{is it possible to automatically and precisely make any video into any style in the user query?}

To answer this question, 
we attempt to observe how video editors work in our realistic world.
We notice that,
video stylization is a complex workflow which typically involves  three important stages:
video shot parsing,
style type selecting,
video style rendering \cite{video_rendering}.
Inspired by this observation,
we think that video stylization requires systematical video understanding and distinct style analyzing,
instead of developing a single generative model.
To this end,
we propose to design a generic AI agent system for video stylization,
\textbf{V-Stylist},
which can effectively render any video into the required style in the open query of user,
by
mimicking video professionals with a collaborative pipeline of Multi-modal Large Language Models (MLLMs).
Specifically,
our V-Stylist consists of three critical roles,
as shown in \cref{fig:teaser}.

(1) \textit{Video Parser}.
To handle complex video transitions,
we introduce a concise \textit{video-to-shot prompting} paradigm.
First,
we leverage a shot detector to decompose the input video into a number of shots,
according to video transitions.
Then,
we employ a shot captioner to describe the key content of each shot.
Finally,
we use a shot translator to convert the detailed caption of each shot into its corresponding prompt,
which is used as textual condition in the generation model to maintain content consistency in the stylization of each shot.
Based on such a top-down design,
video parser can effectively address videos with various transitions,
by stylizing each shot with its content prompt.

(2) \textit{Style Parser}.
To tackle vague style preference,
we introduce a robust \textit{tree-of-thought searching} paradigm.
First,
we leverage Large Language Model (LLM) as a style identifier,
in order to parse the specific style from the open description of user query.
Second,
we need to develop a generation model that is used to transform the video into the required style.
However,
directly training such a model with the style prompt is problematic,
due to no training samples on hand for a general user.
Alternatively,
we notice that,
in the AI community like huggingface\cite{huggingface} and civitai\cite{civitai},
there exist a number of open-source diffusion models that are well-pretrained on various styles.
Hence,
instead of training a model,
we propose to build a style tree for searching the model,
according to dependency of various styles.
Specifically,
we leverage LLMs as experts and progressively decide how to move in the tree via experts discussion.
Such a tree-of-thought searching manner allows to robustly find the model that effectively matches the required style of user query,
alleviating difficulty of vague style preference.

(3) \textit{Style Artist}.
To deal with fixed structure control,
we develop a style artist with a novel multi-round \textit{self-reflection} mechanism on the stylized video.
Given a video shot and its content prompt,
we first leverage the matched model to stylize it.
To control various types of structure details,
we add a set of corresponding ControlNets (e.g. lineart, tile, depth) that are initially assigned with the same weight.
After generating the stylized shot,
we design a style reflection role to check if detail control is satisfactory in this shot.
If unsatisfactory,
we will assign a set of new weights to ControlNets.
Via multi-round rendering and reflection, 
we adaptively adjust detail control for better stylization.

It is worth mentioning that,
our V-Stylist is not a simple combination of multiple models.
Instead,
it leverages an insightful agent collaboration and reflection manner of multi-modal large language models,
which largely alleviates three primary challenges in the existing approaches for versatile and robust video stylization in practice.
Moreover,
since existing benchmarks lack assessment of complex videos on diversified style queries,
we further develop a new benchmark of TVSBench (Text-driven Video Stylization Benchmark),
which consists of 50 videos with multiple scene transitions, and 17 styles for assessment.
Extensive experiments validate the superiority of our V-Stylist on TVSBench, 
where it consistently outperforms state-of-the-art competitors,
e.g., V-Stylist outperforms FRESCO and ControlVideo in overall metrics by 6.05\% and 4.51\% respectively, 
indicating its superior capability of capturing the stylistic essence for stylizing complex videos from open user queries.
The code and model is all available at \href{https://zhengrongyue.github.io/v-stylist.github.io/}{https://V-Stylist.github.io}.


%% file: sec/2_related_work.tex
\section{Related Work}
\label{sec:formatting}

\begin{figure*}[t]
    \centering
    \includegraphics[width=\textwidth]{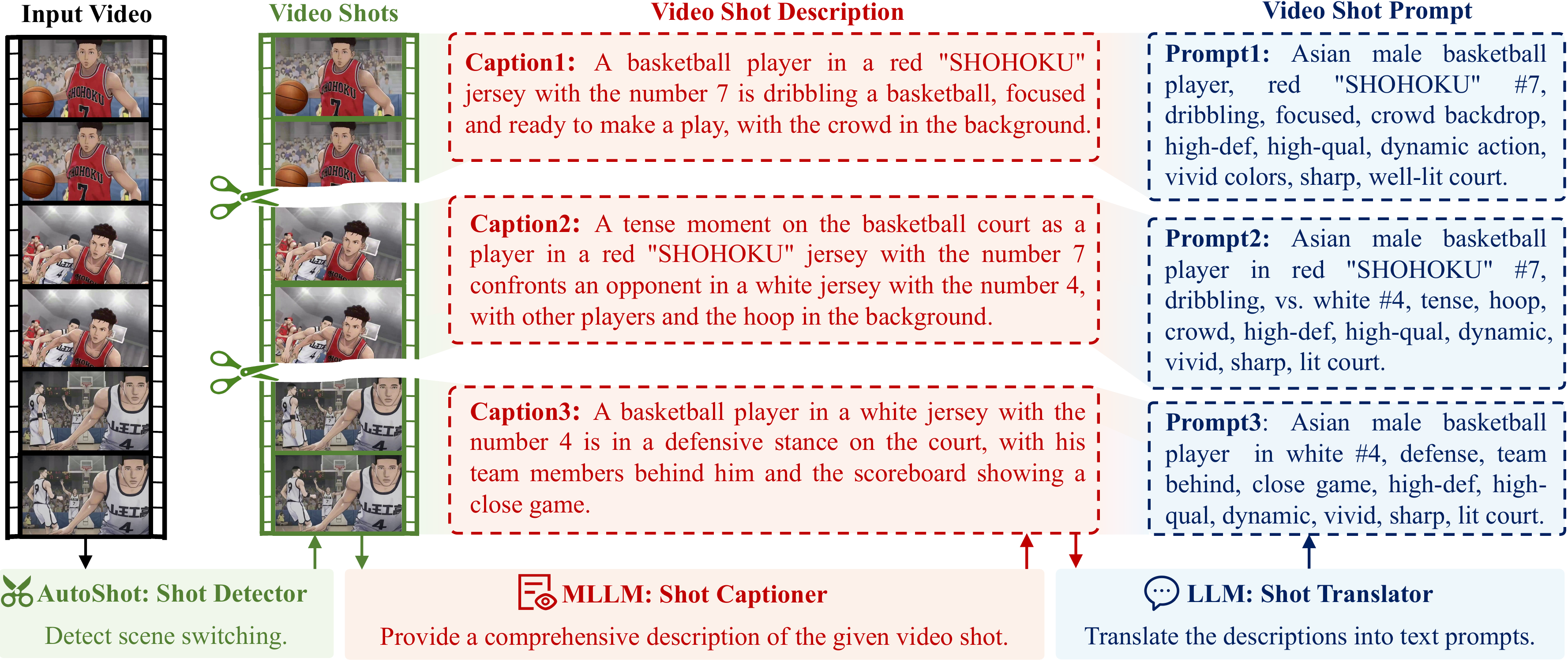}
    \caption{\textbf{Video Parser.} 
    First, 
    Shot Detector splits video into a number of shots,
    based on transitions. 
    Then, 
    Shot Captioner generates caption to describe key content of each shot.
    Finally, 
    Shot Translator converts shot captions into text prompts for diffusion later on.}
    \label{fig:VideoParser}
    \vspace{-0.3cm}
\end{figure*}

\noindent \textbf{Video Stylization.}
In recent years, 
video stylization has gained significant attention. 
Early methods \cite{coherent, arbitrary, fast, artistic, interactive, exploring, Learning} applied image stylization techniques \cite{Gatys,AdaIN,WCT,StyTr2} to video frames, 
generating stable stylized video sequences by statistical alignment and optical flow constraints. 
However, 
these approaches are limited,
as certain artistic styles are difficult to fully characterize by the reference image. 
Advancements in pre-trained text-to-image (T2I) models \cite{stable-diffusion, imagen, dalle2} have led to text-driven video stylization methods \cite{gen1, vid2vid-zero, video-p2p, tokenflow, pix2video, fatezero, lovecon, controlvideo, control-a-video, fairy, flowvid, flatten}, 
yet creating prompts that accurately convey complex artistic styles remains challenging.
Additionally, 
model-driven methods \cite{rerender, fresco, diffutoon} pre-trained models on various style domains to stylize target videos but require users to manually select suitable style models, 
making it a labor-intensive process.
These methods also struggle with style-specific structure control, as a single control strategy \cite{sdedit, controlnet} often fails to maintain structural integrity and versatility across diverse style editing tasks. 
To address the limitations above, 
we build an innovative V-Stylist framework by collaboration and reflection of multi-modal large language models.

\begin{figure*}[t]
    \centering   \includegraphics[width=\textwidth]{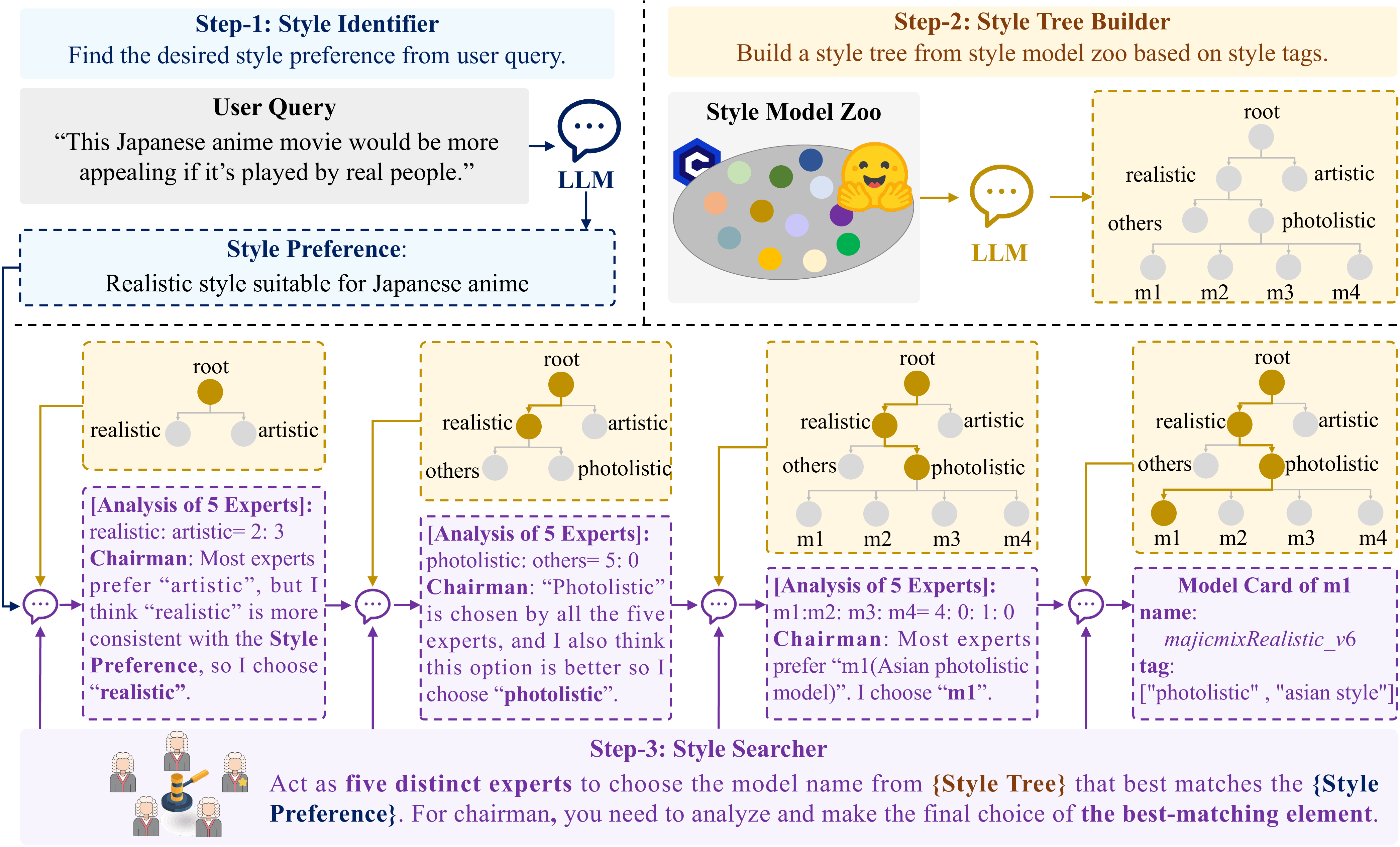}
    \caption{\textbf{Style Parser.} 
    First,
    Style Identifier finds the style preference from open user query.
    Second,
    Style Tree Builder constructs a style tree based on dependency of various styles.
    Finally,
    Style Searcher uses tree-of-thought to search the matched style model.}
    \label{fig:StyleParser}
    \vspace{-0.3cm}
\end{figure*}

\noindent \textbf{Multi-modal Agent System.}
Powered by LLMs \cite{llama2, llama3, gpt4} and MLLMs \cite{llava, qwen2vl, internvl},
multi-agent system combine text and vision capabilities to execute complex multi-modal tasks across various domains.
Applications range from visual understanding \citep{visiongpt, visualgpt} and vision-language navigation \cite{vlnagent} to gaming \citep{tom, mindagent}, image generation \cite{visiongpt, genartist, muses,diffusiongpt}, video generation \citep{mora,worldgpt,vlogger}, and autonomous driving \citep{chatsim, multi}. Especially in visual generation, GenArtist \cite{genartist} employs MLLMs for planning tree inference, achieving cohesive image generation and editing. DiffusionGPT \cite{diffusiongpt} uses LLMs for model selection. MUSES \cite{muses} employs LLMs for 3D planning to ensure precise control over visual contents. Vlogger \cite{vlogger} uses LLMs to craft scripts, establishing a workflow for long-form video production. Mora \cite{mora} integrates various image and video generation models into a multi-agent framework. WorldGPT \cite{worldgpt} optimizes prompts with LLMs, combining multiple generative and perceptual models for video generation and editing.
Drawing inspiration from these advancements, our V-Stylist leverages the planning and reflection capabilities of MLLMs to perform complex video stylization with minimal supervision. 

%% file: sec/3_method.tex
\section{Method}

In this section,
we explain in detail the three key components of V-Stylist in \cref{fig:teaser}.
First,
we present Video Parser to deal with complex video transitions. 
Second,
we introduce Style Parser to tackle vague style preference.
Finally,
we develop Style Artist to iteratively refine the stylization process through multi-round reflection on detail control.

\subsection{Video Parser: Video-to-Shot Prompting}
\label{subsec:video_parser}

As mentioned before,
videos in practice often contain complex scene transitions.
However,
most existing approaches are designed for stylizing video clips with only a few seconds.
As a result,
they are limited to handle videos with diversified transitions.
To tackle this challenge,
we propose a concise Video Parser in a video-to-shot prompting paradigm.
It consists of three core agents in \cref{fig:VideoParser},

\textbf{Shot Detector.}
First,
we employ a shot detector (e.g., AutoShot \cite{autoshot}) to detect video transitions.
It can effectively cut the input video into a number of shots $\{\mathcal{X}_{t}\}_{t=1}^{T}$. 

\textbf{Shot Captioner.}
Next,
we leverage MLLM (e.g., Qwen2-VL \cite{qwen2vl}) as a shot captioner, 
in order to generate caption of each shot that describes its key visual content (such as objects, actions, colors, etc) for stylization.

\textbf{Shot Translator.}
Finally,
we use LLM (e.g., Mistral8x7B \cite{mixtral}) as a shot translator, 
for converting the caption of each shot into the text prompt $\{\mathcal{P}_{t}\}_{t=1}^{T}$ used in the diffusion model.
It ensures that the shot content can be correctly understood and applied by the stylization model.

Via such a top-down manner,
Video Parser allows our V-Stylist to make stylization in the shot level,
which can effectively alleviate difficulty in direct stylizing the entire video with multiple transitions.

\subsection{Style Parser: Tree-of-Thought Searching}
\label{subsec:style_parser}

After obtaining video shots with prompts,
the next question is how to precisely capture the required style of an open user query.
To tackle this problem,
we introduce a robust Style Parser that can effectively identify style preference and progressively select the stylization model via tree-of-thought searching.
It consists of three core agents in \cref{fig:StyleParser}.

\textbf{Style Identifier.} 
For a general user,
its query is often open even ambiguous.
Hence,
we start with exploiting the true style requirement from an in-the-wild query.
Specifically,
we leverage LLM (e.g., Mistral8x7B \cite{mixtral}) as style identifier to find style preference $\mathcal{S}$ from user query, 
due to the great linguistic understanding power of LLM.

\begin{figure*}[t]
    \centering  
    \includegraphics[width=\textwidth]{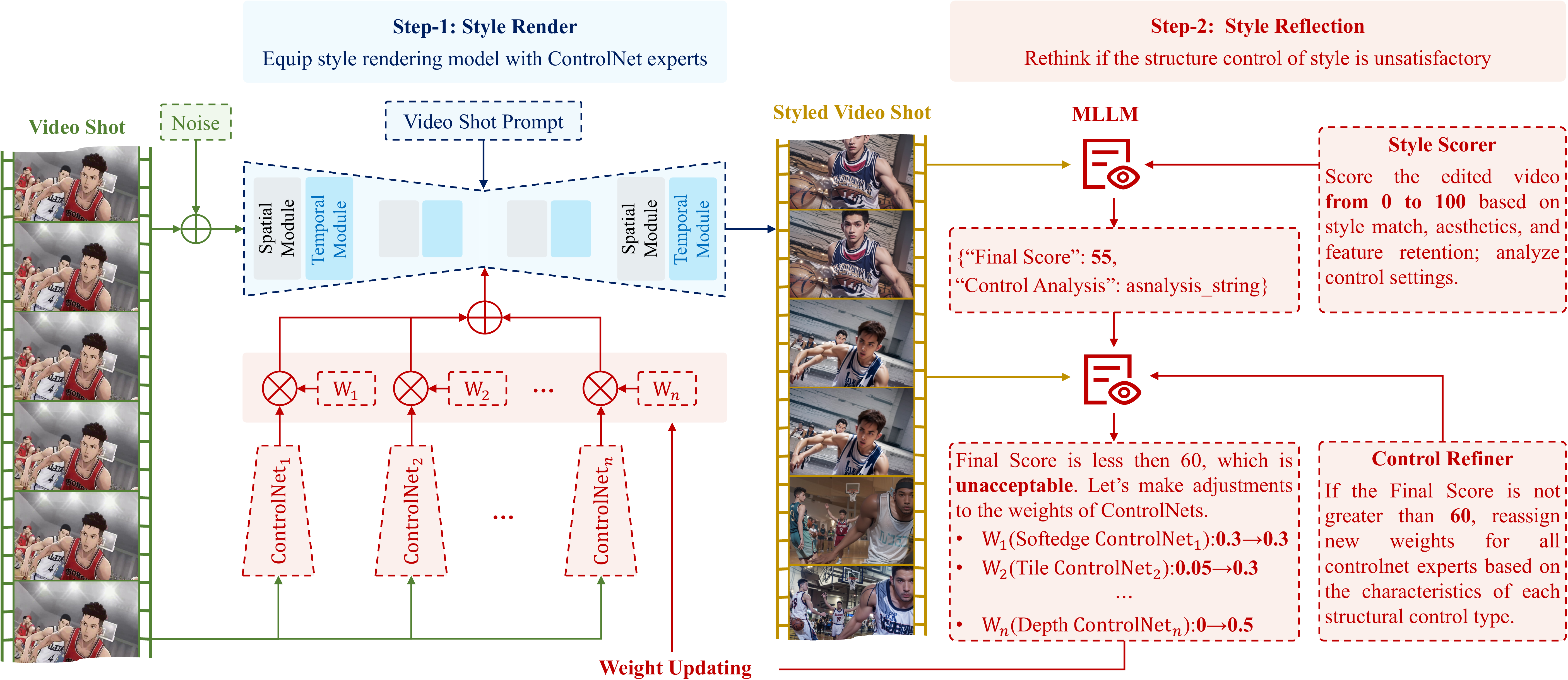}
    \caption{\textbf{Style Artist.} 
    In the Style Render,
    we leverage the matched style model to convert a video shot into the required style,
    based on the current control weights.
    In the Style Reflection,
    we use a Style Scorer to evaluate if the stylized shot is satisfactory.
    If not,
    we use a Control Refiner to generate new control weights for stylization in the next round.
    The two steps iterate alternately, for progressively and adaptively enhancing visual details of stylization.}
    \vspace{-0.1cm}
    \label{fig:VideoStyleArtist}
    \vspace{-0.4cm}
\end{figure*}

\textbf{Style Tree Builder.}
After obtaining the style preference,
the next job is to develop a model for stylizing video shots.
The tradition manner is to train a diffusion model with the given style.
However,
this solution is not practical,
since a general user may only have very limited training samples on hand,
especially for the particular styles in the art.
Alternatively, 
we notice that huggingface community offers numerous open-source diffusion models pre-trained on various styles. 
Hence,
we propose model searching instead of model training for zero-shot need of a general user.
Inspired by \cite{diffusiongpt}, 
we build a style tree according to dependency of various styles to find a matched model systematically. 
First, 
we assemble the metadata of these open-source models,
such as style tags of visual aesthetics, artistic expressions, etc.
Second, 
we leverage LLM to analyze such metadata to identify key style categories and their subcategories.
Finally, 
we construct the style tree by organizing models into these predefined groups, 
ensuring a logical structure for style retrieval. 
Notably, 
this tree is designed with dynamic extensibility, 
allowing for seamless integration of new style models. 

\textbf{Style Searcher.} 
After building the style tree,
the last question is to find the most matched style from this tree.
Inspired by Natural Language Processing \cite{tot, llm-tot},
we design a robust tree-of-thought searching paradigm in this work.
Specifically,
we use the query style $\mathcal{S}$ as reference,
and iteratively search the matched one in the style tree $\mathcal{T}$,
\begin{equation}
\mathcal{D}_{l+1}=\text{LLM}(\mathcal{D}_{l}~|~\mathcal{S}, \mathcal{T}),
\label{eq:tot}
\end{equation}
where 
$\mathcal{D}_{l}$ is the chosen style in the $l$-th level of the style tree,
e.g.,
$l=1$ refers to the root node in the $1$st level.
At the $l$-th searching iteration,
we use LLM to decide which style in the sub-categories of $\mathcal{D}_{l}$ is preferable to match the required style $\mathcal{S}$.
To increase search robustness,
we ask LLM to act as a number of style experts and make decisions on expert meeting.
As shown in \cref{fig:StyleParser},
we set five experts for style searching, and the chairman for making the final decision based on the five style choices. 
The detailed instructions can be found in Supp.Doc.
Finally,
we find the matched style in the bottom level of the tree $\mathcal{D}_{L}$,
which has a corresponding model $\mathcal{M}_{L}$ for stylization.

Via such a tree-of-thought searching manner,
Style Parser allows our V-Stylist to precisely identify the style from an open query,
and progressively find a matched and well-pretrained model from the hierarchical tree of various styles.
As a result,
it can tackle difficulty in vague style preference of a general user for robust video stylization.

\subsection{Style Artist: Rendering with Self-Reflection}
\label{subsec:video_style_Artist}

Via Video Parser and Style Parser,
we respectively obtain video shots and the style model.
The final stage is to leverage the style model to render all video shots into the required style.
To achieve this goal,
we develop a Style Artist in Fig. \ref{fig:VideoStyleArtist},
consisting of style rendering and style reflection.

\textbf{Style Render.}
In this phase,
we stylize a video shot $\mathcal{X}_{t}$ by the style model $\mathcal{M}_{L}$ searched from Style Parser.
As shown in Fig. \ref{fig:VideoStyleArtist},
it is like the standard diffusion process, 
with the input of this video shot $\mathcal{X}_{t}$ and its content prompt $\mathcal{P}_{t}$ from Video Parser.
Additionally,
we notice that the open-source style models are often Text-to-Image (T2I) diffusion models.
Hence,
we equip the T2I model with training-free temporal layer (e.g., AnimateDiff \cite{animatediff}) to maintain temporal consistency of the stylized shot.
Moreover,
we introduce a number of ControlNets $\mathcal{C}_{1:N}$ to control visual details (e.g., softedge, tile, depth, etc.) in the stylized shot.
To weight their importance,
we equip them with weight scores $\mathcal{W}_{1:N}$.
Then,
we summarize the weighted feature as condition of the style model $\mathcal{M}_{L}$,
when transferring the original shot $\mathcal{X}_{t}$ into the stylized one $\mathcal{Y}_{t}$,
\begin{equation}
\mathcal{Y}_{t}=\mathcal{M}_{L}(\mathcal{X}_{t}, \mathcal{P}_{t}~|~\mathcal{C}_{1:N}.\mathcal{W}_{1:N}),
\label{eq:vr}
\end{equation}

\begin{figure*}[tp]
    \centering
    \vspace{-0.2cm}
    \includegraphics[width=\linewidth]{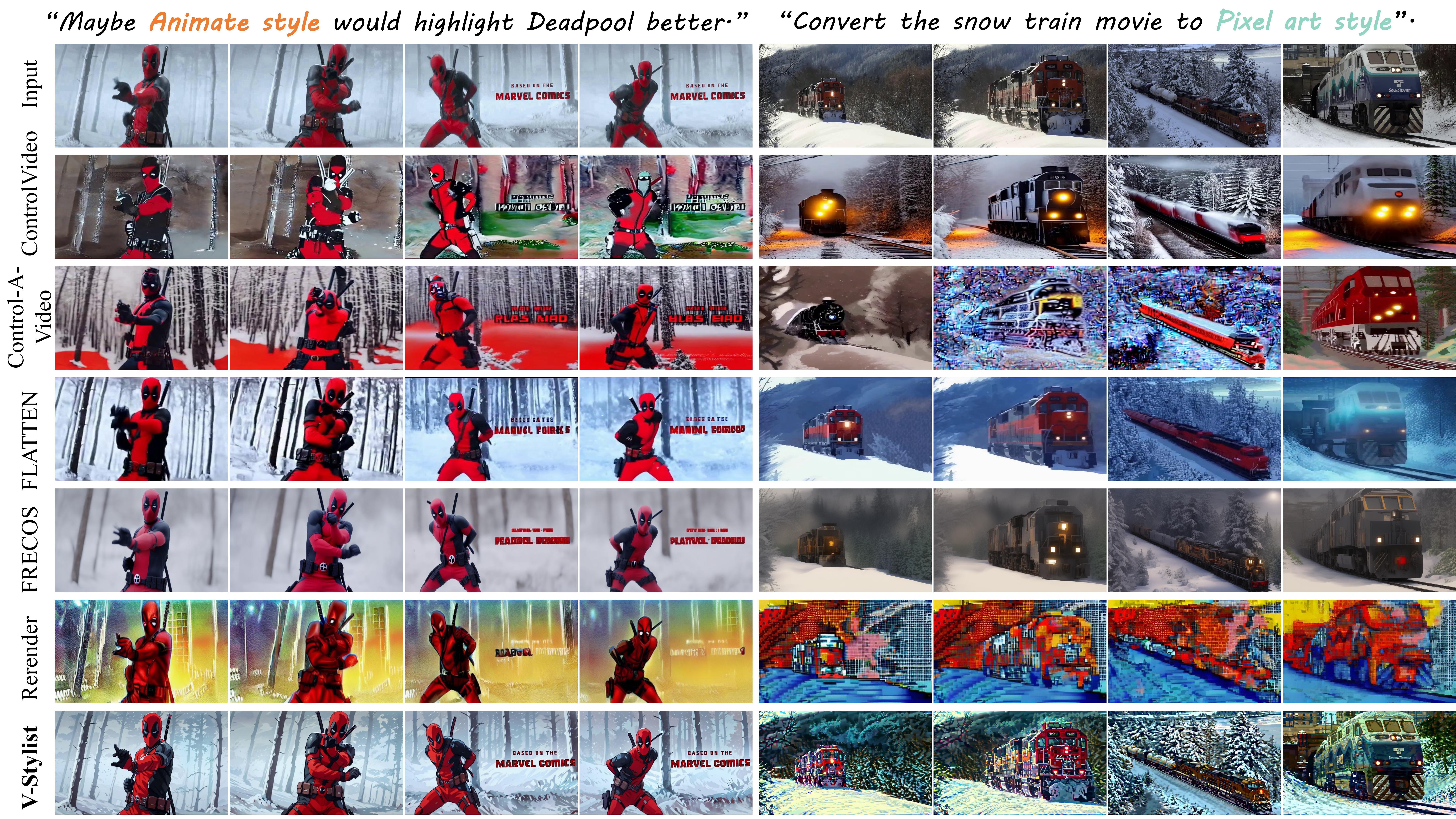}
    \vspace{-0.5cm}
    \caption{\textbf{Qualitative Comparison with State-of-the-Art Methods.} Our V-Stylist achieves the best in terms of condition alignment, temporal consistency, and video quality, outperforming open-source state-of-the-art methods, such as ControlVideo \cite{controlvideo}, Rerender-A-Video \cite{rerender}, and FRESCO \cite{fresco}.}
    \label{fig:SOTA_1}
    \vspace{-0.3cm}
\end{figure*}

\begin{table*}[t]
    \centering
    \setlength{\tabcolsep}{4pt}
    \fontsize{9pt}{11pt}\selectfont
    \resizebox{\textwidth}{!}{
        \begin{tabular}{ll cc cc cccc c}
        \toprule
        
        \multicolumn{2}{l}{\multirow{2}{*}{\textbf{Models}}} & \multicolumn{2}{c}{\textbf{Condition Alignment}} & \multicolumn{2}{c}{\textbf{Temporal Consistency}} & \multicolumn{4}{c}{\textbf{Video Quality}} & \textbf{Overall} \\
        
        \cmidrule(r){3-4} \cmidrule(r){5-6} \cmidrule(r){7-10} \cmidrule(lr){11-11}
        
        & & \textbf{CLIP-T} $\uparrow$ & \textbf{CLIP-W} $\uparrow$ & \textbf{Structure} $\uparrow$ & \textbf{Semantics} $\uparrow$ & \textbf{Aesthetic-I} $\uparrow$ & \textbf{Aesthetic-V} $\uparrow$ & \textbf{Distortion-I} $\uparrow$ & \textbf{Distortion-V} $\uparrow$ & \textbf{Average} $\uparrow$ \\
        
        \midrule
        ControlVideo \cite{controlvideo} & & 0.2631 & {0.1570} & 0.8900 & 0.9733 & 0.5874 & 0.4490 & 0.5868 & 0.5413 & 0.5560 \\
        Control-A-Video \cite{control-a-video} & & 0.2044 & 0.1512 & 0.8102 & 0.9725 & 0.4529 & 0.4761 & 0.5829 & 0.4995 & 0.5187 \\
        FLATTEN \cite{flatten} & & 0.2433 & 0.1504 & 0.6083 & 0.9717 & 0.5284 & 0.4170 & 0.5571 & 0.4197 & 0.4870 \\
        Rerender \cite{rerender} & & 0.2062 & 0.1537 & 0.8499 & 0.9715 & 0.4117 & 0.5541 & 0.4038 & 0.4664 & 0.5022 \\
        FRESCO \cite{fresco} & & 0.2387 & 0.1384 & 0.8322 & 0.9762 & 0.5269 & 0.4844 & 0.5561 & 0.5710 & 0.5405 \\
        
        \midrule
        
        \rowcolor{gray!20} 
        \textbf{V-Stylist (Ours)} & & {0.2669} & 0.1528 & {0.9020} & {0.9772} & {0.5906} & {0.5826} & {0.5924} & {0.7445} & \textbf{0.6011} \\
        
        \bottomrule
        \end{tabular}
    }
    \vspace{-0.3cm}
    \caption{\textbf{Quantitative Comparison with State-of-the-Art Methods on TVSBench.} Our V-Stylist demonstrates the best performance on condition alignment, temporal consistency, and video quality, outperforming SOTA methods.}
    \label{table: comparison}
\vspace{-0.4cm}
\end{table*}

\begin{figure*}[tp]
    \centering
    \includegraphics[width=\linewidth]{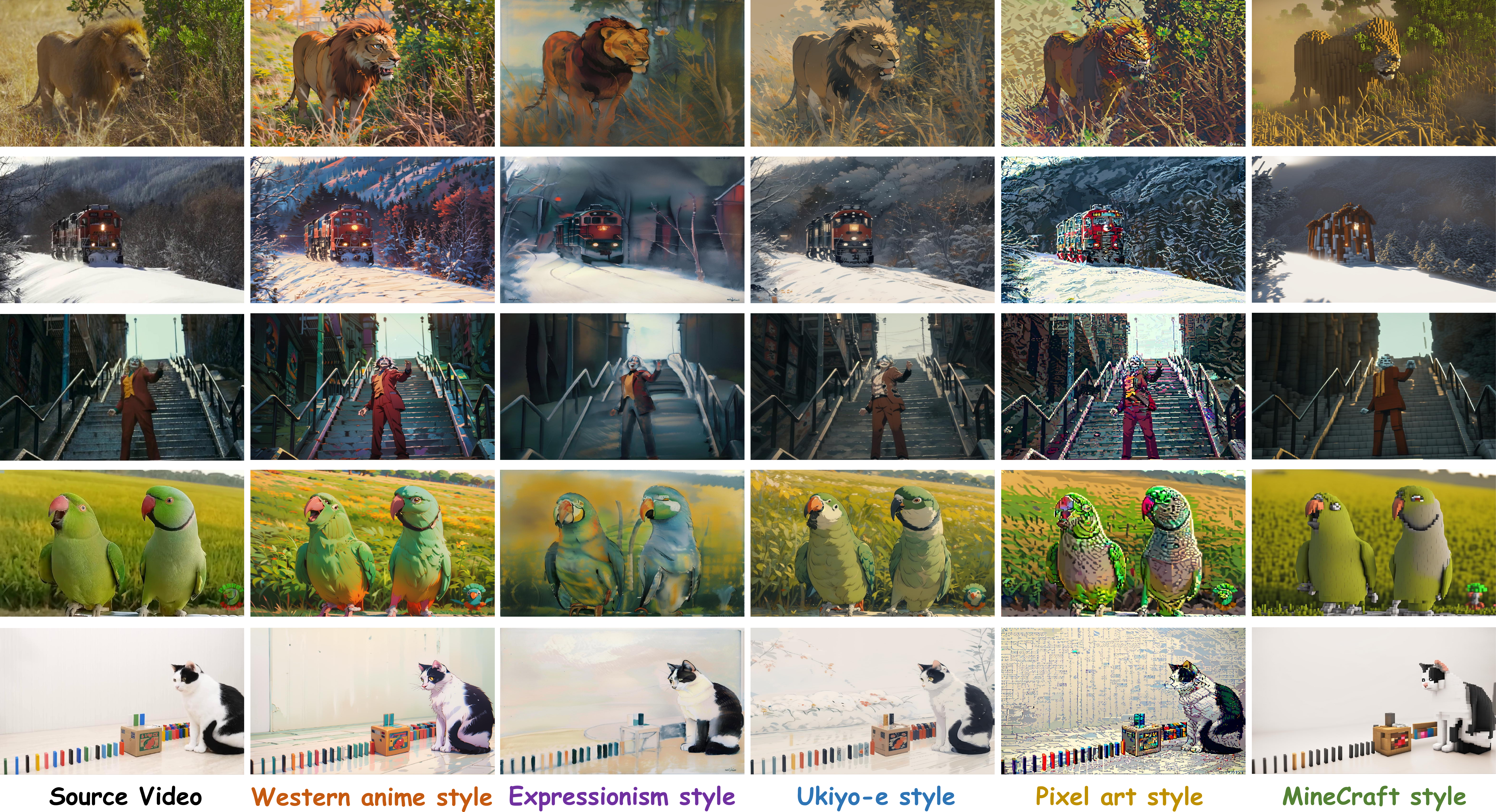}
    \vspace{-0.7cm}
    \caption{
    \textbf{Visualization of V-Stylist's diverse stylization results.} Each row corresponds to the same frame from a video, with the first column displaying the source video and subsequent columns representing specific styles. This demonstrates the powerful style rendering capabilities of V-Stylist.
    }
    \label{fig:Image_Case}
    \vspace{-0.3cm}
\end{figure*}

\begin{table*}[t]
    \centering
    \belowrulesep=0pt\aboverulesep=0pt
    
    \resizebox{\textwidth}{!}{
        \begin{tabular}{ccc|cc cc cccc}
            \toprule

            \multirow{2}{*}{\textbf{VP}} & \multirow{2}{*}{\textbf{SP}} & \multirow{2}{*}{\textbf{SA}} & \multicolumn{2}{c}{\textbf{Condition Alignment}} & \multicolumn{2}{c}{\textbf{Temporal Consistency}} & \multicolumn{4}{c}{\textbf{Video Quality}} \\ 
            
            \cmidrule(r){4-5} \cmidrule(r){6-7} \cmidrule(r){8-11}

            & & & \textbf{CLIP-T} $\uparrow$ & \textbf{CLIP-W} $\uparrow$ & \textbf{Structure} $\uparrow$ & \textbf{Semantics} $\uparrow$ & \textbf{Aesthetic-I} $\uparrow$ & \textbf{Aesthetic-V} $\uparrow$ & \textbf{Distortion-I} $\uparrow$ & \textbf{Distortion-V} $\uparrow$ \\
            
            \midrule

            \textcolor[RGB]{192, 192, 192}{\ding{56}} & \textcolor[RGB]{192, 192, 192}{\ding{56}} & \textcolor[RGB]{192, 192, 192}{\ding{56}} & 
            0.2556 & 0.1248 & 0.8612 & 0.9820 & 0.5569 & 0.6294 & 0.5756 & 0.6204  \\
            
            \ding{51} & \textcolor[RGB]{192, 192, 192}{\ding{56}} & \textcolor[RGB]{192, 192, 192}{\ding{56}} 
            & 0.2627 & 0.1166 & 0.8988 & 0.9838 & 0.5687 & 0.6473 & 0.5844 & 0.6364  \\
            
            \ding{51} & \ding{51} & \textcolor[RGB]{192, 192, 192}{\ding{56}} 
            & 0.2655 & 0.1459 & 0.8849 & 0.9863 & 0.5912 & 0.6509 & 0.5749 & 0.6630  \\
            
            \ding{51} & \ding{51} & \ding{51} 
            & \textbf{0.2662} & \textbf{0.1519} & \textbf{0.9041} & \textbf{0.9867} & \textbf{0.5950} & \textbf{0.6887} & \textbf{0.5895} & \textbf{0.7028}  \\
            
            \bottomrule
        \end{tabular}
    }        
    \vspace{-0.2cm}
    \caption{\textbf{Ablation Study on TVSBench-highlight.} We evaluate the results for various components of the system. Without the Video Parser (VP), only style words are utilized. Without the Style Parser (SA), models are chosen randomly. Without the Style Artist (SA), control reflection optimization is not executed, and inference relies solely on random initial control scales.}
    \vspace{-0.3cm}
    \label{table: ablation}
\vspace{-0.3cm}
\end{table*}

\textbf{Style Reflection.}
A natural question is how to assign the values for these weights $\mathcal{W}_{1:N}$,
since different styles require to control different visual details.
Apparently,
the manual attempt is sub-optimal,
due to diversified video content and open style requirement.
Hence,
we propose a novel multi-round self-reflection phrase to adaptively adjust the weight scores,
by verifying if the stylized video shot is satisfactory.
Suppose that,
the weights at the $i$-th round are $\mathcal{W}_{1:N}^{(i)}$,
where
we set the same value for all the weights at the initial round $i=1$.
Based on $\mathcal{W}_{1:N}^{(i)}$,
we can obtain the stylized shot at this round $\mathcal{Y}_{t}^{(i)}$,
according to Eq. (\ref{eq:vr}).
Then,
we leverage MLLM (e.g., Qwen2-VL \cite{qwen2vl}) as a style scorer to evaluate the style of $\mathcal{Y}_{t}^{(i)}$ with a score (from 0 to 100),
\begin{equation}
\mathcal{R}^{(i)}=\text{MLLM}(\mathcal{Y}_{t}^{(i)}),
\label{eq:ss}
\end{equation}
according to style match, aesthetics, etc.
If the score $\mathcal{R}^{(i)}$ is higher than threshold (e.g., 60),
we believe $\mathcal{Y}_{t}^{(i)}$ at round $i$ is satisfactory and use it as the final stylized shot.
If the score is lower than threshold,
we believe $\mathcal{Y}_{t}^{(i)}$ is not acceptable.
As a result,
we leverage the same MLLM as the control refiner,
and generate new weights for ControlNets,
\begin{equation}
\mathcal{W}_{1:N}^{(i+1)}=\text{MLLM}(\mathcal{Y}_{t}^{(i)}, \mathcal{R}^{(i)}),
\label{eq:cr}
\end{equation}
based on the stylized shot $\mathcal{Y}_{t}^{(i)}$ and its style score $\mathcal{R}^{(i)}$.
Then,
we start style rendering and reflection in the next round.
Note that,
we set a largest round to stop such a cycle.
If the stylized shot is still not satisfactory at the largest round,
we stop and choose the round where the style score is the highest.
See detailed MLLM prompts in Supp.Doc.

Via such a multi-round rendering  and style reflection manner,
Style Artist allows our V-Stylist to adaptively adjust visual details,
according to different shot content and visual styles.
As a result,
it alleviates the problem of fixed detail control to enhance video stylization performance.

%% file: sec/4_experiment.tex
\section{Experiment}

\noindent\textbf{Datasets and Metrics.}
In the evolving field of video generation, existing benchmarks show limitations, particularly in assessing video-to-video stylization.
Some concentrate on text-to-video and image-to-video conversions, overlooking video-to-video evaluations \cite{vbench, evalcrafter, aigcbench}. 
Others lack a thorough assessment of long-form videos, which are more challenging \cite{TGVE, ebench, v2vbench}. 
We introduce \textbf{TVSBench}, 
a benchmark designed for video stylization nuances, using a curated dataset of 50 high-quality videos from the internet. 
These videos, averaging 30 seconds at 30 FPS, cover five complex stylization tasks: large motion, small objects, similar foreground and background, occlusion, and multiple object interactions.
For efficient ablation, we offer \textbf{TVSBench-highlight}, a 5-second highlight version of these videos, capturing the most representative segments.
%
The distribution of video categories in the dataset and the types of instructions are detailed in Supp.Doc.
The TVSBench metrics cover \textit{Condition Alignment}, \textit{Temporal Consistency}, and \textit{Video Quality}. \textit{Condition Alignment} checks text-frame congruence and style-frame match. \textit{Temporal Consistency} gauges structural coherence and semantic preservation. \textit{Video Quality} evaluates aesthetics, technical in frame and video level. More details are in the Supp.Doc.

\begin{figure}[t]
    \centering
    \resizebox{0.48\textwidth}{!}{\includegraphics{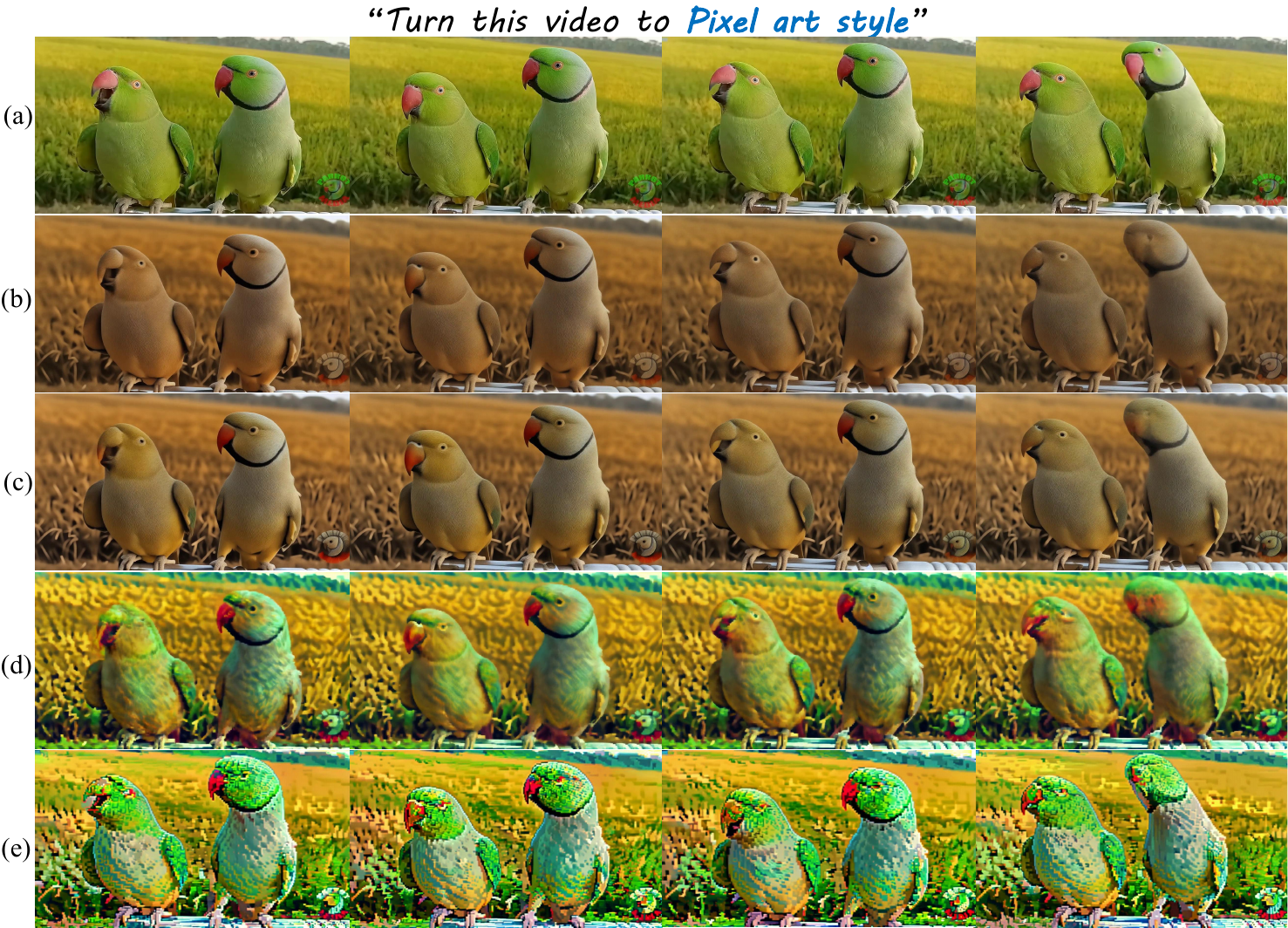}}
    \vspace{-0.7cm}
    \caption{\textbf{Qualitative Results on Ablation Study.} 
    (a) Input Video. 
    (b) Initial approach using vanilla video diffusion model and fixed controlnet. 
    (c) Enhanced with the integration of Video Parser. 
    (d) Further enhanced by incorporating Style Parser.
    (e) V-Stylist, integrating all components including Style Artist.}
    \label{fig:Ablation}
    \vspace{-0.7cm}
\end{figure}

\noindent\textbf{Implementation Details.}
Our V-Stylist is a flexible and extensible framework that seamless integrates MLLMs, Video Diffusion Models, and ControlNets. 
In our experiments, we employed Qwen2-VL \citep{qwen2vl} as our MLLM and Mistral8x7B \citep{mixtral} as our LLM, setting top-k to 10, top-p to 0.95, and temperature to 0.7.
Detailed prompts and in-context examples can be found in Supp.Doc.
We employed AutoShot\cite{autoshot} as our Shot Detector in the Video Parser. For each shot, we extracted 3 frames to generate detailed captions.
The Style Tree in the Style Parser encompasses 17 styles with 25 leaf nodes (models), at a depth of 3. The number of experts is 5 in Style Searcher.
Following Diffutoon \cite{diffutoon}, the Video Diffusion Model in Style Artist consists of Stable Diffusion v1.5 \cite{stable-diffusion}, AnimateDiff \cite{animatediff}, and ControlNet v1.1 \cite{controlnet}. 
In our experiments, we utilized 4 specific ControlNets including tile, depth, softedge, and lineart, with initial weights setting in Supp.Doc.
The Style Reflection process is capped at 3 rounds.
Experiments are all conducted on an Ubuntu 20.04 system with 8 NVIDIA RTX A6000 GPUs.

\noindent\textbf{SOTA Comparison.} 
As presented in \cref{table: comparison}, we compared our V-Stylist with existing SOTA methods, demonstrating superior performance across all evaluated metrics. V-Stylist consistently outperforms both specialized and generic models in Condition Alignment, Temporal Consistency, and Video Quality. 
All methods are based on stable diffusion v1.5.
Specifically,
The integration of our Video Parser and Style Parser strengthens the correlation between video content and text prompts,
while maintaining style alignment, 
achieving the best performance in condition alignment. 
Additionally, 
the Style Parser promotes aesthetic style and reduces distortion at both the frame-level and video-level, 
leading to superior video quality. 
Moreover, our innovative Style Artist employs a sophisticated multi-round control adaptive reflection mechanism, 
enhancing advanced structure analysis capabilities, 
and resulting in top scores for temporal consistency and condition alignment. 
As shown in \cref{fig:SOTA_1}, 
the qualitative comparison confirms that 
V-Stylist yields videos that 
align with user preferences, 
display high visual quality, 
and exhibit excellent temporal consistency.

\begin{figure}[tp]
    \centering  
    \resizebox{0.48\textwidth}{!}{\includegraphics{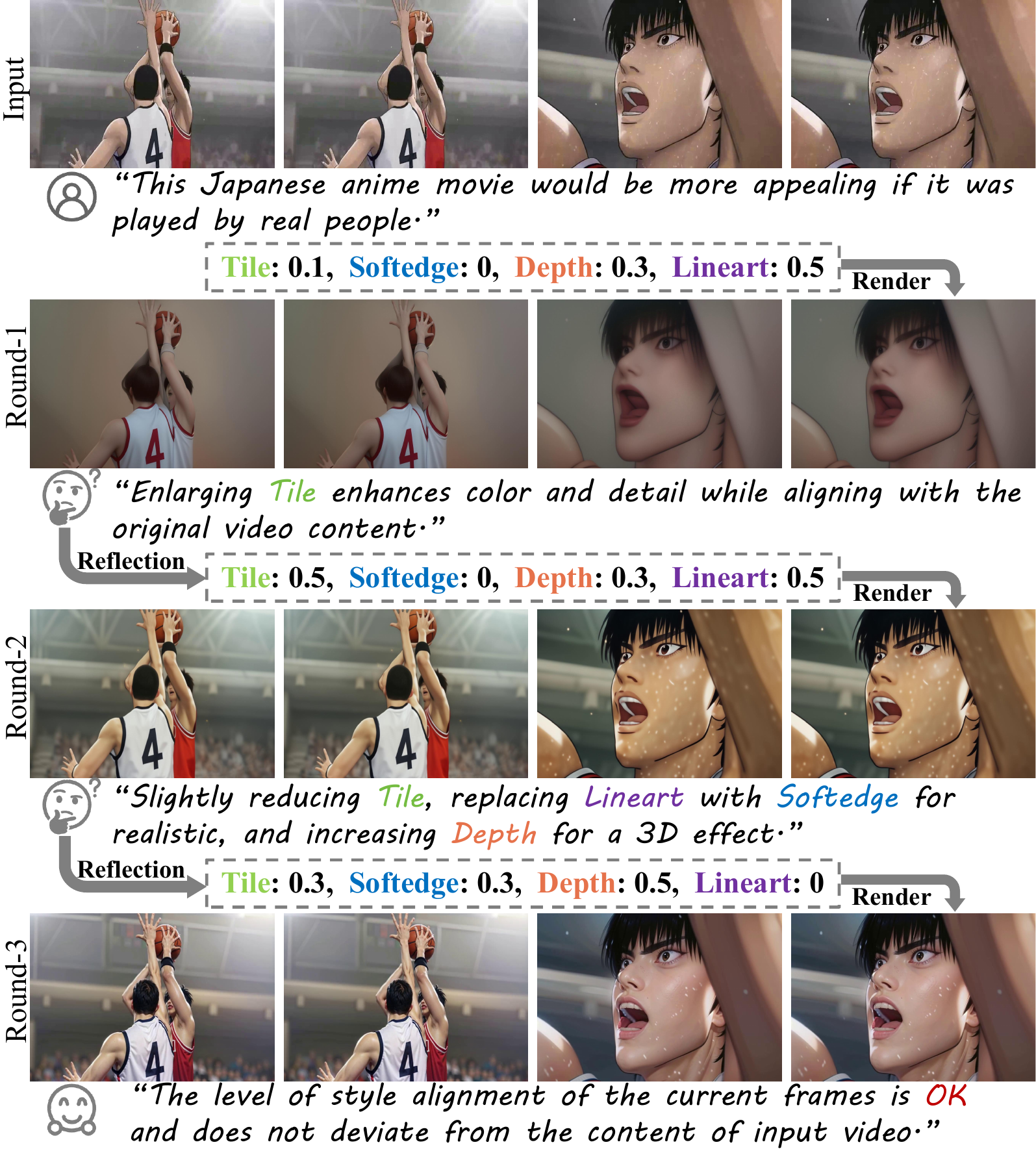}}
    \vspace{-0.7cm}
    \caption{\textbf{Style Reflection.} After three rounds of reflection and optimization, the degree of style alignment, adherence to the original video, and aesthetic quality are continuously improved.}
    \label{fig:Control_Reflection}
    \vspace{-0.7cm}
\end{figure}

\noindent\textbf{Ablation Studies.}
First,
we visualize V-Stylist with various styles.
As shown in \cref{fig:Image_Case},
our V-Stylist can successfully achieve stylization on diversified styles,
indicating its power on practical style rendering.
Second,
we discover that each system component positively affects video generation performance, 
via our ablation study on the TVSBench-highlight of \cref{table: ablation}.  
Adding the Video Parser (VP) enhances text alignment and aesthetics, likely by enriching frame content with improved text prompts, but it slightly reduces style alignment due to a decrease in the proportion of style words in the overall prompt. 
For instance, in \cref{fig:Ablation}, (c) exhibits more vivid colors compared to (b).
Incorporating the Style Artist (SA) significantly improves video quality and style alignment by 25.16\%, underscoring the importance of selecting the appropriate stylized model. 
Compared to (c), (d) shows a noticeable emergence of pixel art style and aesthetic aspect.
The Style Artist (SA) further refines results through iterative reflection optimization, addressing style-specific structural control demands, and enhancing style fidelity. 
Through fine-grained structure control reflection, (e) further elevates the level of stylization, successfully aligning with the user's style preferences.
When all components are integrated, the system achieves its best performance, demonstrating that significance of each component in a collaborative, agent-centric system design.
Finally,
we visualize style reflection to show if it works on detail control.
As shown in \cref{fig:Control_Reflection},
our style reflection can effectively adjust visual details in the stylized shot.
via progressively updating control weights of different conditions. 
More detailed results in Supp.Doc.

%% file: sec/5_conclusion.tex
\section{Conclusion}
In conclusion, we introduce V-Stylist, a multimodal reflection agent system designed for the stylization of long videos.
By leveraging self-reflection to iteratively optimize control parameters, V-Stylist achieves fine-grained control over visual effect.
To evaluate such complex long video scenes more comprehensively,
we construct a new benchmark TVSBench.
Experiments highlight V-Stylist's exceptional ability to manage complex long videos, delivering high-quality stylization. In future work, we plan to optimize system efficiency and broaden our video rendering models, aiming to realize diverse video editing.

\newpage

%% file: sec/6_acknowledgments.tex
\section{Acknowledgments}
This work was supported by the National Key R\&D Program of China (NO.2022ZD0160505), and the Shenzhen Key Laboratory of Computer Vision and Pattern Recognition.



%% file: sec/X_suppl.tex
\appendix
\setcounter{page}{1}
\maketitlesupplementary

\section{TVSBench Details}
\label{sec:tvsbench}

\begin{figure*}[tp]
    \centering
    \includegraphics[width=\textwidth]{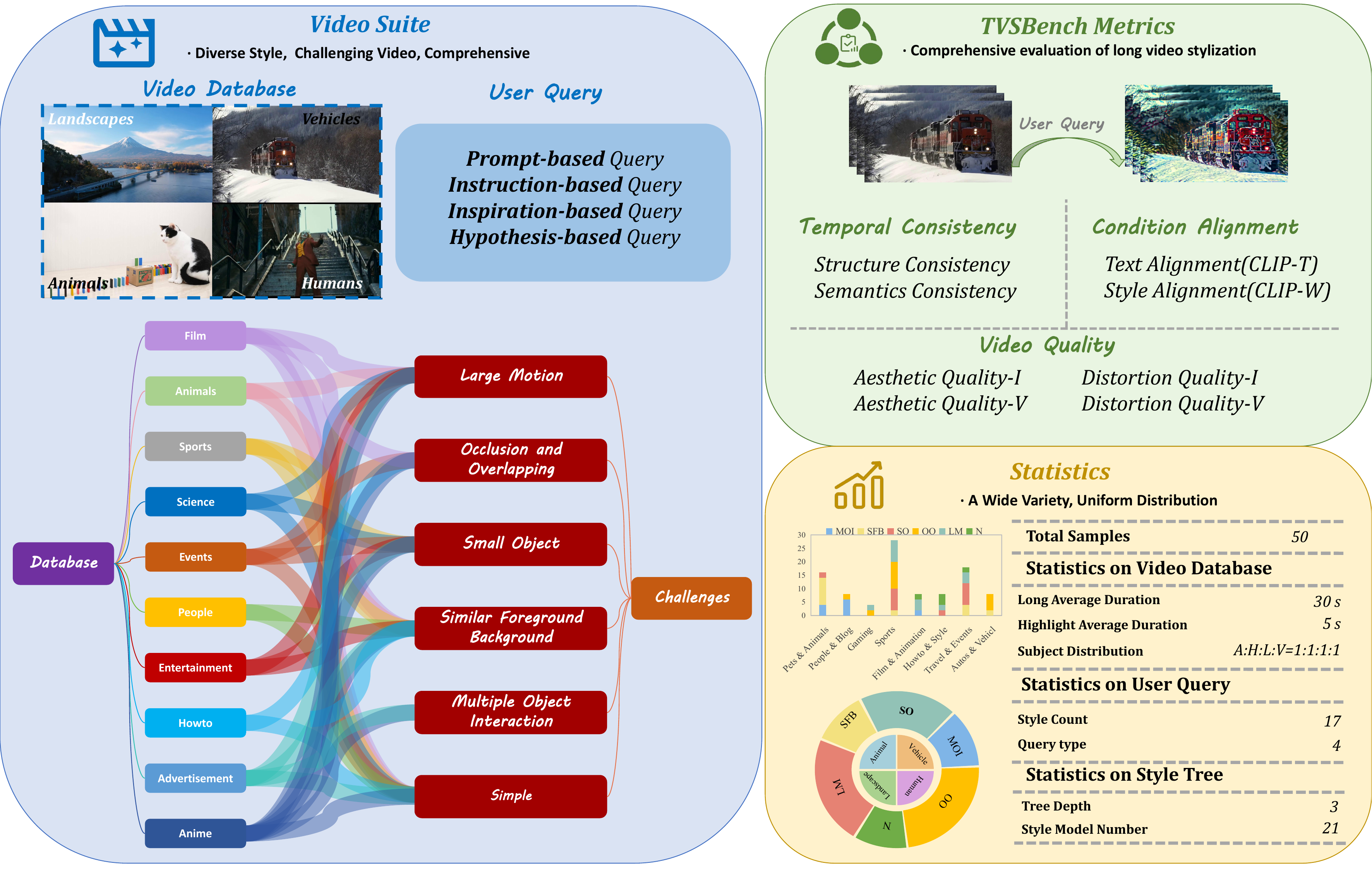}
    \caption{\textbf{Overview of the TVSBench.} The left portion of the figure showcases the extensive video database, encompassing a multitude of genres including landscapes, vehicles, animals, and human subjects. It underscores the complexities involved in stylizing videos with significant motion, instances of occlusion and overlapping, the presence of small objects, challenging foreground and background similarities, intricate multiple object interactions, and the pursuit of stylistic simplicity. Furthermore, it elaborates on the spectrum of open user queries, ranging from those driven by prompts-based, instructions-based, inspirations-based, to hypothesis-based. The upper right section delineates the evaluative metrics integral to TVSBench, which encompass temporal and structural consistency, semantic coherence, condition and text alignment, style congruence, as well as the dual aspects of aesthetic and distortion qualities within video content. The lower right section presents a comprehensive set of statistics. These include a diverse and evenly distributed array of video samples, alongside an in-depth analysis of style counts, the categorization of user queries, the architectural depth of the style tree, and the inventory of style models, collectively providing a robust framework for video stylization assessment.}
    \label{fig:TVSBench}
\end{figure*}

Due to the lack of a suitable complex video stylization dataset that includes open user queries, multiple scene transitions, and challenging cases (e.g., large motion, small object), we have constructed our benchmark, namely TVSBench (Text-driven Video Stylization Benchmark), as depicted in \cref{fig:TVSBench}.

Given the simplicity and static nature of video samples in current academic video editing, which struggle to meet the complexities of real-world video applications, we have collected 50 more challenging video samples targeting five key challenges in video stylization: large motion, occlusion and overlaying, small objects, similar foreground backgrounds, and multiple object interactions. The average duration of the full-length videos is 30 seconds at 30 fps, while the highlight versions average 5 seconds at 30 fps. The videos in the database are categorized into five distinct YouTube genres, with the subjects evenly distributed across Humans, Landscapes, Animals, and Vehicles, with an equal proportion for each category.

Each video is paired with a single text user query indicating the desired style preference. These queries follow four modal patterns established by \cite{diffusiongpt}, and the construction process involves the meticulous creation of four types of open user queries:
\begin{itemize}
  \item \textbf{Prompt-based:} These queries are straightforward requests for a specific artistic style without additional context. They are typically used when the user has a clear vision of the desired outcome and wishes to communicate it directly to the system. e.g., \textit{``Pixel art style.''} This prompt indicates the user wants the video to be stylized in the manner of pixel art, a style characterized by small, block-like images.

  \item \textbf{Inspiration-based:} These queries provide a broader context or a thematic inspiration for the style, often referencing a setting or scenario that embodies the desired aesthetic. They are used when the user wants to convey a mood or atmosphere that aligns with their vision. e.g., \textit{``I would love to see a western realistic style video set in a baseball game.''} This prompt suggests the user is looking for a video that captures the realism of a western genre, specifically within the context of a baseball game.

  \item \textbf{Instruction-based:} These queries are more directive, specifying actions or subjects within the video content that need to be stylized in a particular way. They are used when the user has specific instructions for how elements within the video should be treated stylistically. e.g., \textit{``Render this man who is practicing kung fu in a claymation style.''} This prompt instructs the system to stylize a specific action (a man practicing kung fu) in the style of claymation, a technique that uses models made from clay or other malleable materials.

  \item \textbf{Hypothesis-based:} These queries propose a potential style as a hypothesis, often with a degree of uncertainty or suggestion. They are used when the user is unsure of the best style or is open to suggestions from the system. e.g., \textit{``Perhaps a Japanese anime style is the best choice to enhance this video's aesthetics.''} This prompt hypothesizes that applying a Japanese anime style could improve the video's visual appeal, leaving room for the system to confirm or propose alternatives.
\end{itemize}
To enrich the dataset, we utilized GPT4 \cite{gpt4} to mimic and generate an additional 40 similar texts, which were then refined by human experts to ensure quality, diversity, and consistency.

The TVSBench quantitative metrics span three dimensions: \textit{Condition Alignment}, \textit{Temporal Consistency}, and \textit{Video Quality}. 
Condition Alignment evaluates CLIP-T, measuring CLIP \cite{clip} score between frame content and text prompts, and CLIP-W, assessing the match between style words and frames by CLIP \cite{clip}.
Temporal Consistency measures structural coherence with the SSIM Score \cite{ssim} and semantic preservation with the CLIP Score \cite{clip}. 
Video Quality assesses image-level aesthetics and technical appeal using Aesthetic Quality-I and Distortion Quality-I, and video-level aesthetics by the LAION aesthetic predictor \cite{laion} and MUSIQ image quality predictor \cite{musiq}, while DOVER \cite{dover} provides technical integrity with Aesthetic Quality-V and Distortion Quality-V.

\begin{algorithm}[tp]
\caption{Style Reflection Algorithm}
\begin{algorithmic}[1]
\State Given video shot $\mathcal{X}_{t}$, style model $\mathcal{M}_{L}$, content prompt $\mathcal{P}_{t}$, and ControlNets $\mathcal{C}_{1:N}$ with weights $\mathcal{W}_{1:N}$
\State $\mathcal{Y}_{t} \gets \mathcal{M}_{L}(\mathcal{X}_{t}, \mathcal{P}_{t} \mid \mathcal{C}_{1:N}.\mathcal{W}_{1:N})$ \Comment{Stylize the shot}
\State Initialize weights $\mathcal{W}_{1:N}^{(1)}$ and set $i = 1$
\State $\mathcal{Y}_{t}^{(i)} \gets \mathcal{Y}_{t}$
\State $\mathcal{R}^{(i)} \gets \text{MLLM}(\mathcal{Y}_{t}^{(i)})$ \Comment{Evaluate style}
\While{$\mathcal{R}^{(i)} < 60$ and $i \leq T$}
    \State $\mathcal{W}_{1:N}^{(i+1)} \gets \text{MLLM}(\mathcal{Y}_{t}^{(i)}, \mathcal{R}^{(i)})$ \Comment{Refine weights}
    \State $\mathcal{Y}_{t}^{(i+1)} \gets \mathcal{M}_{L}(\mathcal{X}_{t}, \mathcal{P}_{t} \mid \mathcal{C}_{1:N}.\mathcal{W}_{1:N}^{(i+1)})$ \Comment{Re-stylize the video shot}
    \State $\mathcal{R}^{(i+1)} \gets \text{MLLM}(\mathcal{Y}_{t}^{(i+1)})$
    \If{$\mathcal{R}^{(i+1)} > \mathcal{R}^{(i)}$}
        \State $\mathcal{R}^{(i)} \gets \mathcal{R}^{(i+1)}$
        \State $\mathcal{W}_{1:N}^{(i)} \gets \mathcal{W}_{1:N}^{(i+1)}$
    \EndIf
    \State $i \gets i + 1$
\EndWhile
\State Output the final stylized shot $\mathcal{Y}_{t}^{(i)}$ with the highest score
\end{algorithmic}
\end{algorithm}

\section{Implementation details of Style Reflection}
\label{sec:style_reflection}

In the paper, the Style Reflection Algorithm is meticulously designed to optimize the weights of ControlNet for precise style transformation in videos. The process commences with the initialization of weights for softedge, tile, depth, and lineart, all set to a random value between 0.1 and 0.3. This initial setup lays the groundwork for subsequent reflection processes, where a human preference example for each style category serves as in-context guidance for the MLLM to facilitate the reflection process.

The algorithm accepts a video shot $\mathcal{X}_{t}$, along with a style model $\mathcal{M}_{L}$, a content prompt $\mathcal{P}_{t}$ derived from the Video Parser, and a set of ControlNets $\mathcal{C}_{1:N}$ with their respective weights $\mathcal{W}_{1:N}$. The video shot is then stylized by applying the style model $\mathcal{M}_{L}$, which incorporates the content prompt $\mathcal{P}_{t}$ and the weighted ControlNets to produce the stylized shot $\mathcal{Y}_{t}$. The stylized shot $\mathcal{Y}_{t}$ is then evaluated by an MLLM, which assigns a style score $\mathcal{R}^{(i)}$ based on the style match, aesthetics, and other criteria. This score serves as a measure of how well the stylized shot meets the desired style criteria. If the score is below the threshold of 60, indicating that the stylized shot is not satisfactory, the algorithm enters the Style Reflection phase.

In the Style Reflection phase, the algorithm adaptively adjusts the weight scores $\mathcal{W}_{1:N}$ through a multi-round self-reflection process. The weights at the $i$-th round are denoted as $\mathcal{W}_{1:N}^{(i)}$, and initially, all weights are set to the same value. Based on these weights, a new stylized shot $\mathcal{Y}_{t}^{(i)}$ is generated, and its style is evaluated by the MLLM, resulting in a new style score $\mathcal{R}^{(i)}$.

If the style score $\mathcal{R}^{(i)}$ is higher than the threshold, the stylized shot $\mathcal{Y}{t}^{(i)}$ is considered satisfactory and is used as the final output. If not, the MLLM is used again to generate new weights $\mathcal{W}_{1:N}^{(i+1)}$, taking into account the previous stylized shot $\mathcal{Y}_{t}^{(i)}$ and its style score $\mathcal{R}^{(i)}$. This initiates another round of style rendering and reflection.

The algorithm sets a maximum number of rounds $T$ to prevent an infinite loop. If the stylized shot remains unsatisfactory after the maximum number of rounds, the algorithm terminates and selects the round with the highest style score as the final output. This approach ensures that the V-Stylist can dynamically adjust to different video content and style requirements, enhancing the flexibility and effectiveness of the video stylization process.

Through this iterative process, the Style Reflection Algorithm allows for a more nuanced and adaptive control over the visual details of the stylized video shots, moving beyond a one-size-fits-all approach to style transformation. This results in a more personalized and higher-quality stylization that aligns with the diverse and complex nature of video content and user preferences.

\section{Style Tree Details}
\label{sec:style_tree}

\begin{figure*}[tp]
    \centering
    \includegraphics[width=\textwidth]{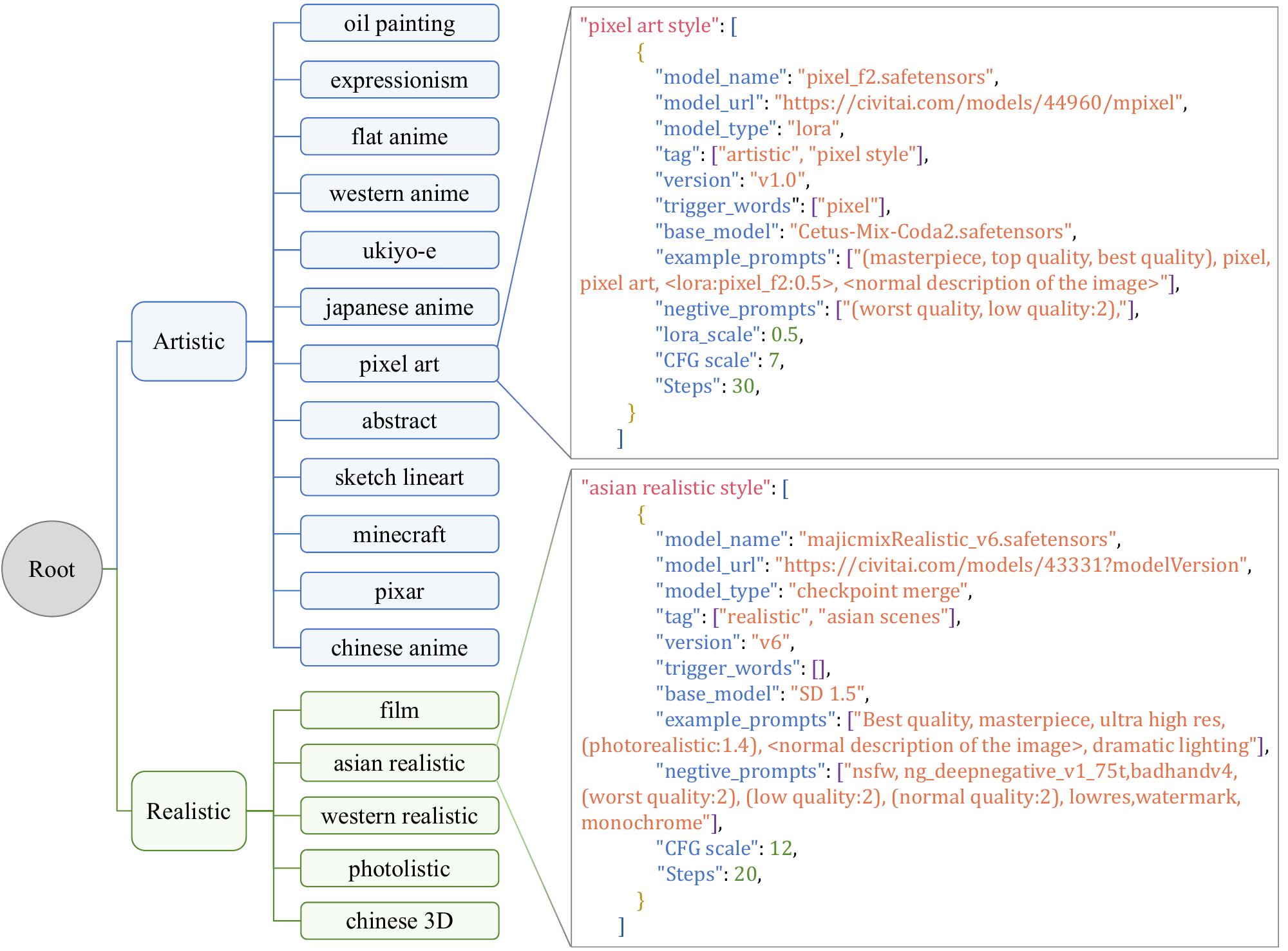}
    \caption{\textbf{Style Tree.} The Style Tree categorizes a variety of artistic and realistic styles within a hierarchical framework. It extends into branches that represent the subcategories falling under the broader categories of Artistic and Realistic styles. The entire tree is organized in JSON format, where each leaf node corresponds to a model card for a specific stylization model. These model cards include details such as model names, URLs, types, tags, and other parameters that are instrumental in directing the stylization process.}
    \label{fig:Style_Tree}
\end{figure*}

When constructing the style tree, we gathered a number of models of 17 various styles and their corresponding model cards from CivitAI \cite{civitai}. These models were categorized into two major classes: Artistic and Realistic. Each distinct model was then mapped to the appropriate style category as leaf nodes within the branches of the tree, as illustrated in the \cref{fig:Style_Tree}. This systematic approach allowed us to create a comprehensive and organized structure that represents the diversity of styles available, making it easier to navigate and select the desired style for specific applications or projects. Naturally, this tree is designed to be dynamically scalable, accommodating additional styles as they are developed.

Under the Artistic, we find styles that lean towards creative expression and abstract representation, such as oil painting, expressionism, and various forms of anime, including flat anime, western anime, and japanese anime. This category also includes unique styles like ukiyo-e, pixel art, and abstract art, each with its own set of models and characteristics. For instance, the ``pixel art style" is an example within this category, with the model ``pixel\_f2.safetensors" that is tagged with ``artistic" and ``pixel style" and is triggered by the keyword ``pixel".

The Realistic category, on the other hand, encompasses styles that aim to replicate or enhance the appearance of real-world visuals. This includes styles like asian realistic, western realistic, and photolistic, etc. For the ``asian realistic style," we have the model ``majicmixRealistic\_v6.safetensors" which falls under the realistic category, specifically tailored to capture the essence of Asian scenes. This model is a checkpoint merge type, indicating that it combines the capabilities of multiple models to produce highly realistic outputs. It is tagged with ``realistic" and ``asian scenes," and it operates on version ``v6" of the base model ``SD 1.5."

Each distinct model within the Artistic and Realistic categories is mapped as a leaf node within the corresponding branch of the Style Tree. This hierarchical structure not only aids in navigation but also provides a clear overview of the relationships between different styles. As illustrated in \cref{fig:Style_Tree}, the tree is designed to be dynamically scalable, allowing for the incorporation of new styles and models as they emerge, ensuring that the Style Tree remains a comprehensive resource for style-based applications and projects.

 \begin{figure}[!htpb]
    \centering
    \includegraphics[width=\linewidth]{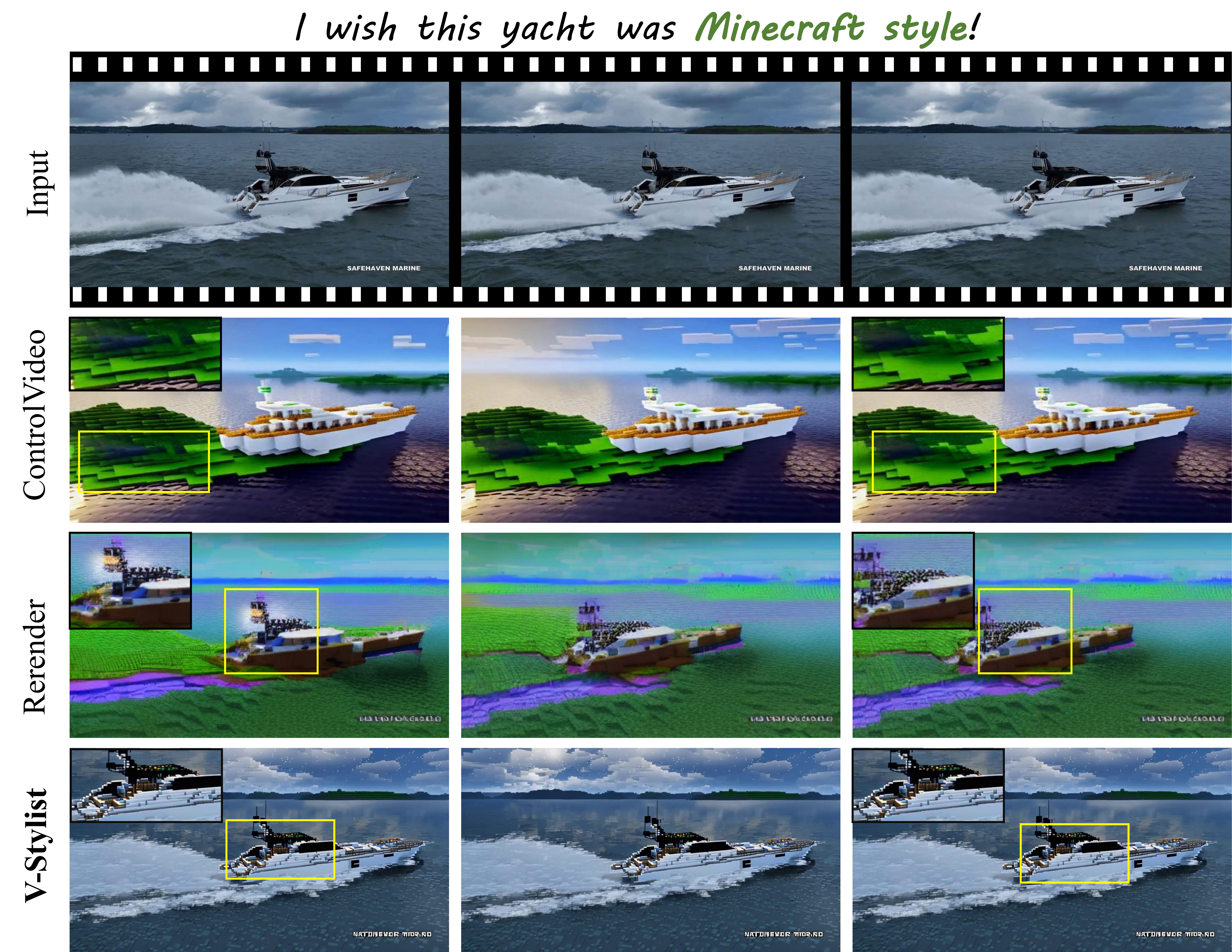}
    \caption{\textbf{Qualitative Comparison with ControlVideo and Rerender-A-Video.} It can be observed that both ControlVideo and Rerender-A-Video have rendered the yacht's wake as green square land, whereas V-Stylist accurately rendered the wake as waves. Although the CLIP calculates a higher Style Alignment for them, their aesthetic quality of style, as well as the precision of transformation and temporal consistency, are inferior to ours.}
    \label{fig:SOTA_2}
\end{figure}

\section{More Visualization of V-Stylist on Long Video Stylization}
\label{sec:video_vis} 

\begin{figure*}[tp]
    \centering
    \resizebox{\textwidth}{0.95\textheight}{\includegraphics{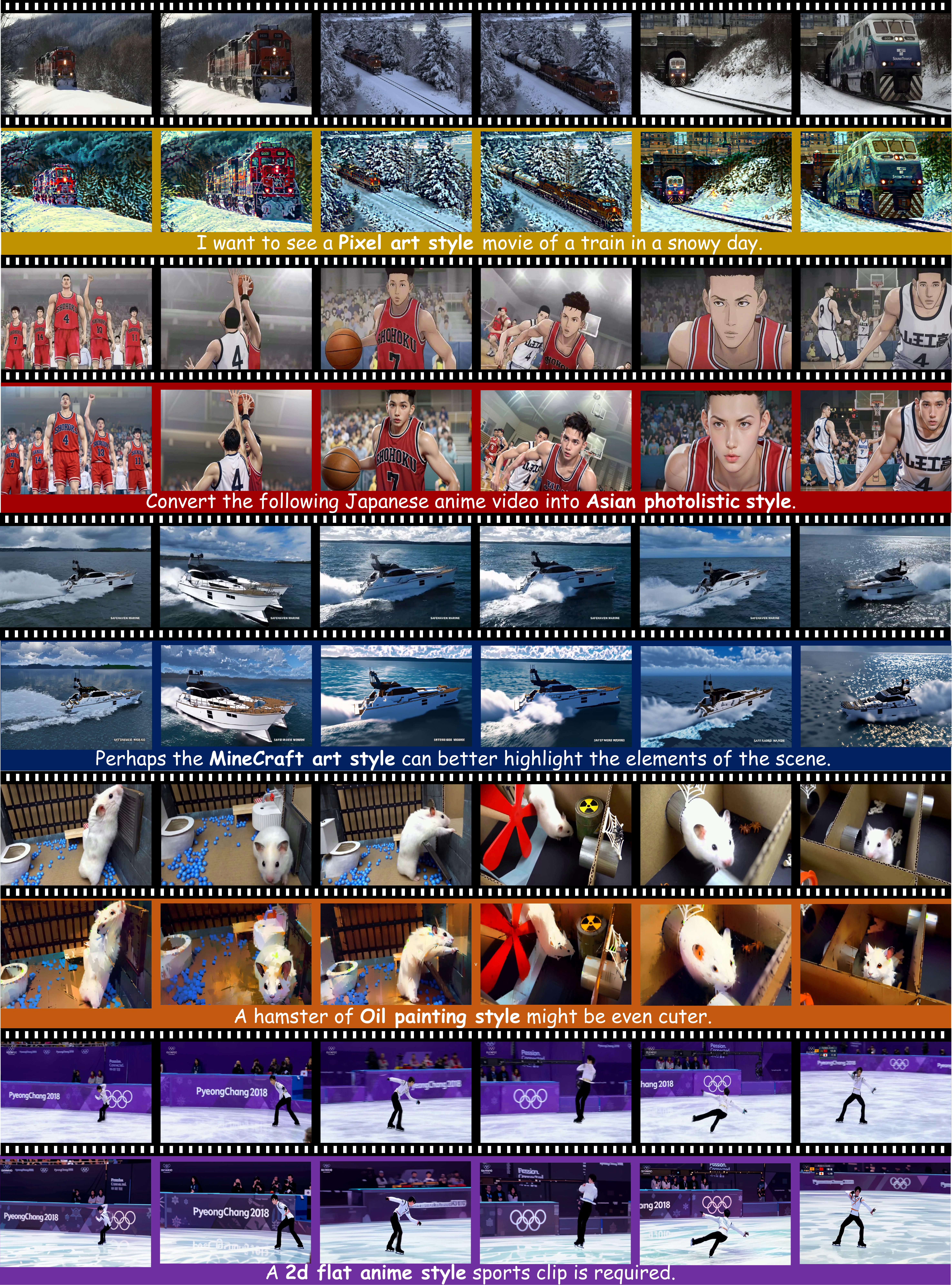}}
    \caption{\textbf{Visualization of different video stylizations of V-Stylist.}}
    \label{fig:examples_1}
\end{figure*}

\cref{fig:examples_1} demonstrates the qualitative results of our V-Stylist on longer video stylization. Our systematic collaborative approach yields videos with high condition alignment, temporal consistency, and video quality.

\section{More Qualitative Comparisons}
\label{sec:qualitative_comparisons}

\begin{figure*}[tp]
    \centering
    \includegraphics[width=\textwidth]{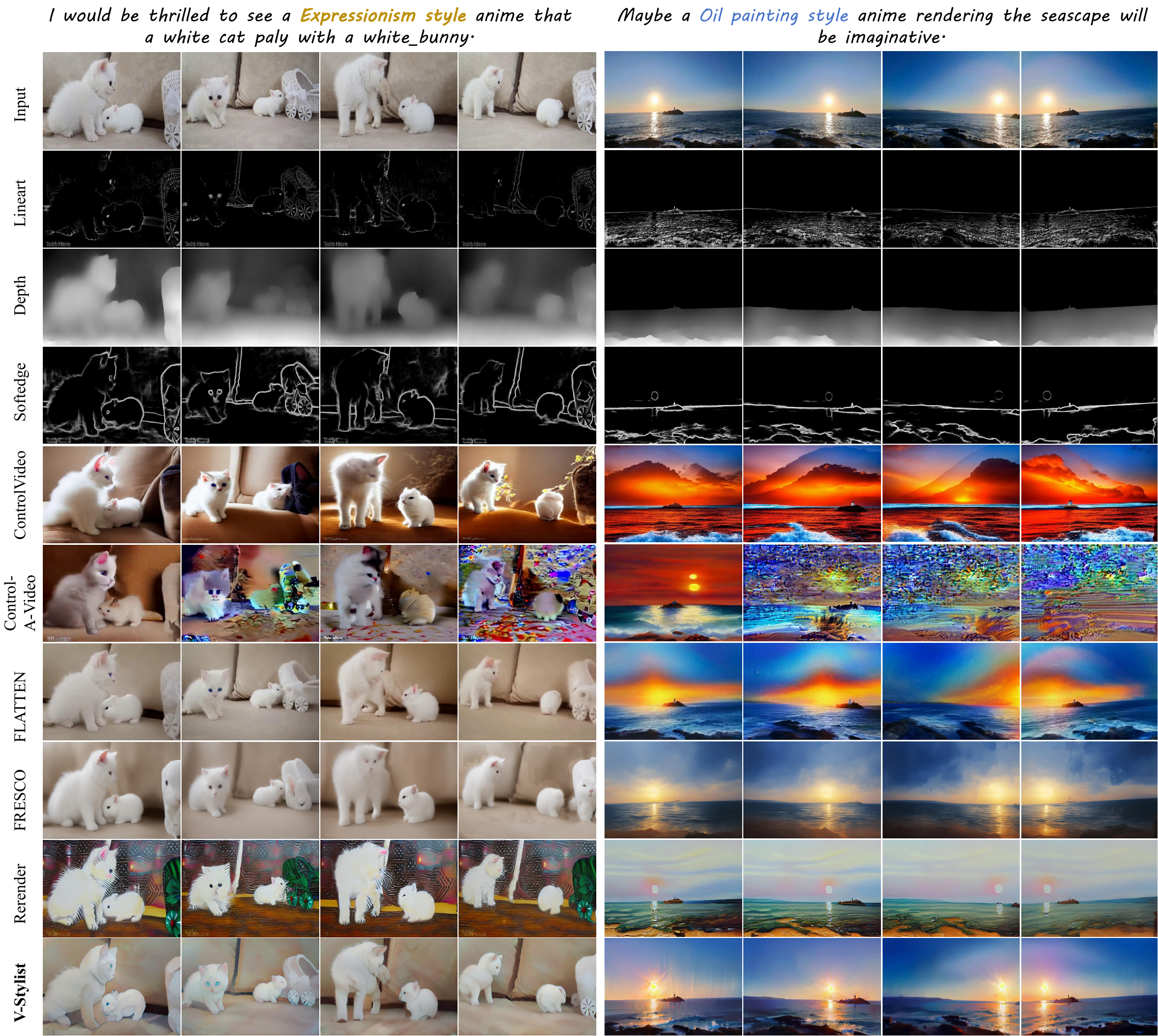}
    \caption{\textbf{Qualitative Comparison (1) With Existing SOTA Methods.} Our V-Stylist consistently demonstrates the highest condition alignment, temporal consistency, and video quality. We compare our V-Stylist with SOTA open-sourced models, including Rerender-A-Video, FRESCO, ControlVideo, FLATTEN, Control-A-Video. The first row is the original video frame, the second to third rows are different structure control images, the fourth to ninth rows are results from different SOTA methods, and the last row is the result from V-Stylist. Color-marked texts at the top indicate specific style preferences.}
    \label{fig:SOTA3-1}
\end{figure*}

\begin{figure*}[tp]
    \centering
    \includegraphics[width=\textwidth]{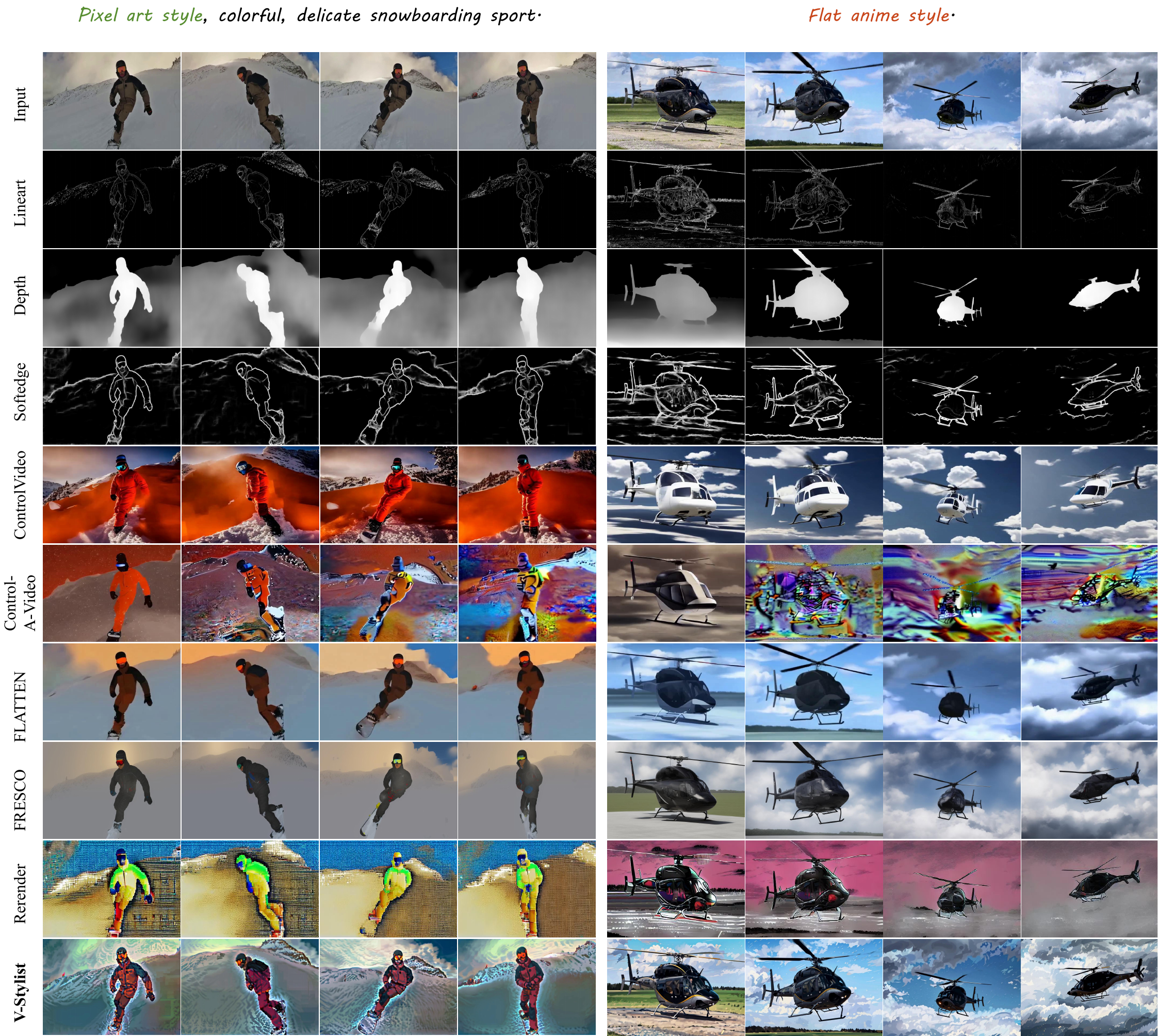}
    \caption{\textbf{Qualitative Comparison (2) With Existing SOTA Methods.} Our V-Stylist consistently demonstrates the highest condition alignment, temporal consistency, and video quality. We compare our V-Stylist with SOTA open-sourced models, including Rerender-A-Video, FRESCO, ControlVideo, FLATTEN, Control-A-Video. The first row is the original video frame, the second to third rows are different structure control images, the fourth to ninth rows are results from different SOTA methods, and the last row is the result from V-Stylist. Color-marked texts at the top indicate specific style preferences.}
    \label{fig:SOTA3-2}
\end{figure*}

\begin{figure*}[tp]
    \centering
    \includegraphics[width=\textwidth]{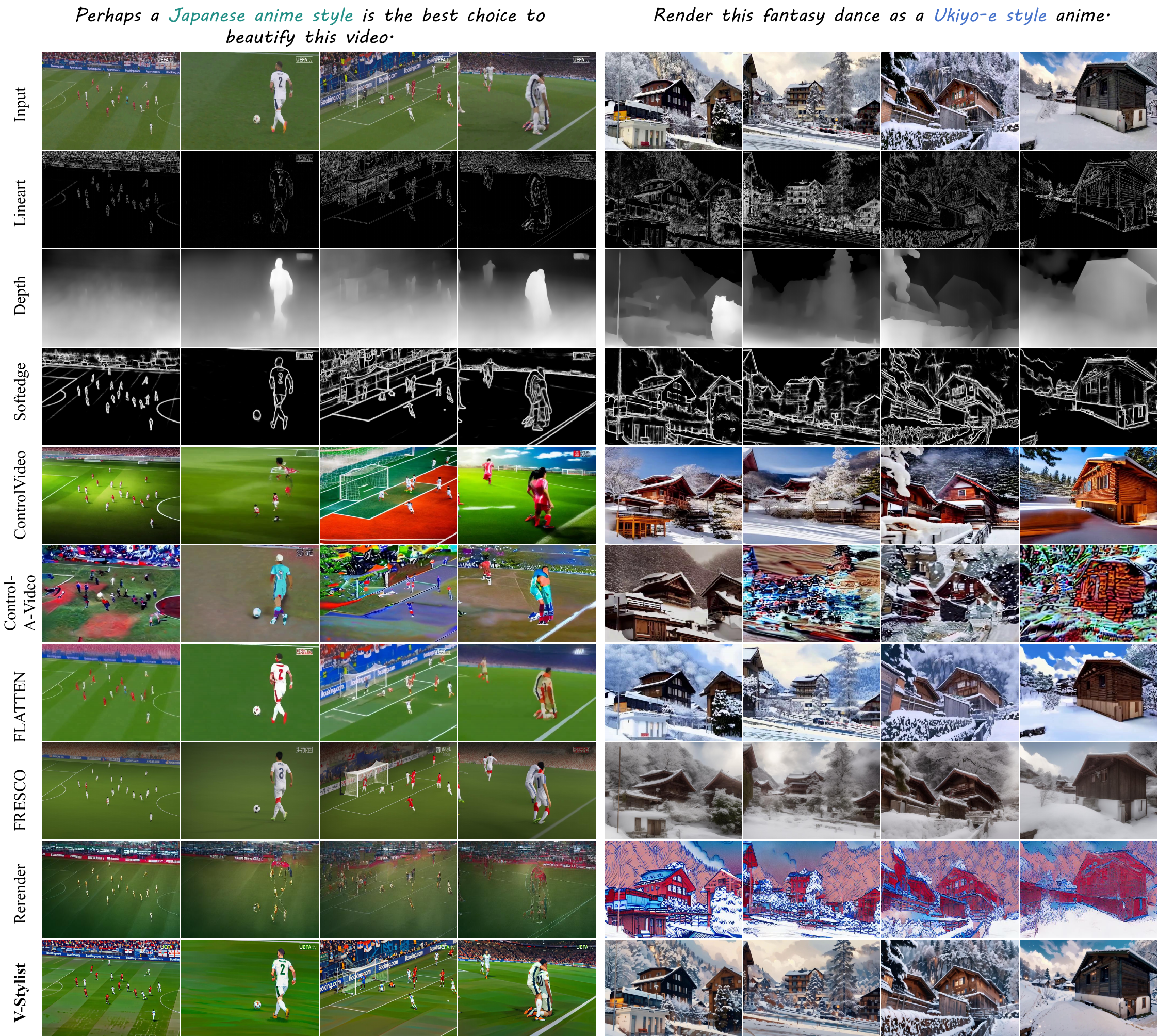}
    \caption{\textbf{Qualitative Comparison (3) With Existing SOTA Methods.} Our V-Stylist consistently demonstrates the highest condition alignment, temporal consistency, and video quality. We compare our V-Stylist with SOTA open-sourced models, including Rerender-A-Video, FRESCO, ControlVideo, FLATTEN, Control-A-Video. The first row is the original video frame, the second to third rows are different structure control images, the fourth to ninth rows are results from different SOTA methods, and the last row is the result from V-Stylist. Color-marked texts at the top indicate specific style preferences.}
    \label{fig:SOTA3-3}
\end{figure*}

\cref{fig:SOTA3-1}, \cref{fig:SOTA3-2}, and \cref{fig:SOTA3-3} respectively display the stylized results of different SOTA methods under various styles. The second to fourth rows are the condition images for Lineart, Depth, and Softedge, while the condition images for Tile are the original images. It can be observed that Control-A-Video \cite{control-a-video} exhibits a noticeable color degradation phenomenon in long videos, while other methods, due to their use of a single ControlNet \cite{controlnet} and rigid weight settings, cannot balance the structure and color information as well as the alignment of styles. Our V-Stylist, by dynamically combining different types of structure controls, consistently demonstrates the highest condition alignment, temporal consistency, and video quality. Moreover, for some styles that are difficult to describe through text or are scarce in the training data, relying solely on text-driven rendering of the base model is challenging to achieve effective results.
Different methods may employ various models. For uniformity, we use the original version of Stable Diffusion v1.5 \cite{stable-diffusion} for Rerender-A-Video \cite{rerender}, FRESCO \cite{fresco}, ControlVideo \cite{controlvideo}, FLATTEN \cite{flatten}, and Control-A-Video \cite{control-a-video}. For structure control, we utilize the default settings for each method: Rerender-A-Video employs ControlNet-HED, FRESCO uses ControlNet-Depth and DDIM Inversion, FLATTEN uses DDIM Inversion, ControlVideo uses ControlNet-Depth, and Control-A-Video uses ControlNet-Depth.

Although the CLIP-W score of V-Stylist is slightly lower compared to Rerender-A-Video and ControlVideo, its actual performance is by no means inferior. 
This is evident when we examine \cref{fig:SOTA_2}, where both ControlVideo and Rerender-A-Video incorrectly render the yacht's wake as green square land, a characteristic of the Minecraft style but a misrepresentation of the original video's content. However, V-Stylist not only applies the Minecraft style effectively but also preserves the original content by accurately rendering the wake as waves, aligning with the natural depiction of a yacht's movement on water. V-Stylist also excels in temporal consistency, ensuring that the style transformation is smooth and coherent across frames, which is vital for the viewer's experience in video content. While Rerender-A-Video and ControlVideo may achieve a higher Style Alignment score according to CLIP, their aesthetic quality and fidelity to the original video's content are compromised. V-Stylist's holistic approach to style transfer considers not only the aesthetic style but also the importance of content alignment and temporal consistency, making it a superior choice for high-quality video transformations that respect the original video's essence.
Incorporating other examples in \cref{fig:SOTA3-1}, \cref{fig:SOTA3-2}, and \cref{fig:SOTA3-3}, we hypothesize that the lower CLIP metric scores for V-Stylist imply that CLIP might not be fully equipped to capture the subtleties of styles, such as Expressionism and other abstract art styles, that are not adequately represented in the training data for CLIP. This insight will also inspire us to further optimize our benchmark metrics to better evaluate the performance on diverse and less conventional artistic styles.

\section{More Ablation Study}
\label{sec:ablation}

Ablation Study of Video Parser compares three different methods of generating stabel diffusion prompts: using only style words, combining raw captions with style words, and using prompts with style words. The results, as shown in \cref{table:video_parser_ablation}, demonstrate the incremental improvements in both text alignment and video quality metrics. The table indicates that the integration of captions with style words and the use of prompts with style words both lead to significant enhancements in video quality metrics compared to using only style words. This suggests that the context provided by captions and prompts is crucial for improving the model's performance in generating high-quality videos that align well with the given conditions.

Ablation Study of Style Parser compares the effectiveness of direct LLM decisions and a progressive LLM search strategy. The results, presented in \cref{table:style_parser_ablation}, highlight the differences in performance between these two approaches. The findings from this study suggest that the progressive search strategy (Tree Search) outperforms both the base model and the direct search method, indicating that a more nuanced approach to style model selection can lead to better alignment and higher video quality. This underscores the importance of a structured search process in achieving optimal style model.

\begin{table}[t]
    \centering
    \setlength{\tabcolsep}{4pt}
    \fontsize{9pt}{11pt}\selectfont
    \resizebox{0.45\textwidth}{!}{
        \begin{tabular}{ll cc ccc}
        \toprule
        
        \multicolumn{2}{l}{\multirow{2}{*}{\textbf{Models}}} & \multicolumn{1}{c}{\textbf{Condition Alignment}} & \multicolumn{3}{c}{\textbf{Video Quality}} \\
        
        \cmidrule(r){3-3} \cmidrule(r){4-7}
        
        & & \textbf{CLIP-T} $\uparrow$ & \textbf{Aesthetic-I} $\uparrow$ & \textbf{Aesthetic-V} $\uparrow$ & \textbf{Distortion-I} $\uparrow$ & \textbf{Distortion-V} $\uparrow$ \\
        
        \midrule

        Only Style Word & & 0.2556 & 0.5569 & 0.6294 & 0.5756 & 0.6204  \\

        Caption + Style Word & & 0.2592 & 0.5628 & 0.6383 & 0.5800 & 0.6284  \\
        
        Prompts + Style Word & & \textbf{0.2627} & \textbf{0.5687} & \textbf{0.6473} & \textbf{0.5844} & \textbf{0.6364}  \\
        
        \bottomrule
        \end{tabular}
    }
    \vspace{-0.2cm}
    \caption{\textbf{Ablation Study of Video Parser.} Experiments conducted based on the Stable Diffusion v1.5-base model demonstrate that Shot Captioner and Shot Translator have achieved improvements in text alignment and video quality through step-by-step optimization of the input text prompts.}
    \label{table:video_parser_ablation}
\end{table}

\begin{table}[t]
    \centering
    \setlength{\tabcolsep}{4pt}
    \fontsize{9pt}{11pt}\selectfont
    \resizebox{0.45\textwidth}{!}{
        \begin{tabular}{ll cc cccc}
        \toprule
        
        \multicolumn{2}{l}{\multirow{2}{*}{\textbf{Models}}} & \multicolumn{2}{c}{\textbf{Condition Alignment}} & \multicolumn{3}{c}{\textbf{Video Quality}} \\
        
        \cmidrule(r){3-4} \cmidrule(r){5-8}
        
        & & \textbf{CLIP-T} $\uparrow$ & \textbf{CLIP-W} $\uparrow$ & \textbf{Aesthetic-I} $\uparrow$ & \textbf{Aesthetic-V} $\uparrow$ & \textbf{Distortion-I} $\uparrow$ & \textbf{Distortion-V} $\uparrow$ \\
        
        \midrule

        Base Model & & 0.2627 & 0.1166 & 0.5687 & 0.6473 & 0.5844 & 0.6364  \\

        Direct Search & & 0.2655 & 0.1300 & 0.5780 & 0.6600 & 0.5900 & 0.6500  \\
        
        Tree Search & & \textbf{0.2662} & \textbf{0.1519} & \textbf{0.5950} & \textbf{0.6887} & \textbf{0.5895} & \textbf{0.7028}  \\
        
        \bottomrule
        \end{tabular}
    }
    \vspace{-0.2cm}
    \caption{\textbf{Ablation Study of Style Parser.} SD1.5 indicates the base model of Stable Diffusion v1.5 without style model search. Direct Search indicates exporting all style model names and tags, conducting a single round of LLM Q\&A to select the style model; if the search fails (LLM's answer is not in the model list, then the base model is used directly). Tree Search indicates the results obtained using the Style Tree progressive search.}
    \label{table:style_parser_ablation}
\end{table}

\section{Full Prompts for V-Stylist}
\label{sec:prompts}

\begin{figure*}[tp]
    \centering
    \includegraphics[width=\textwidth]{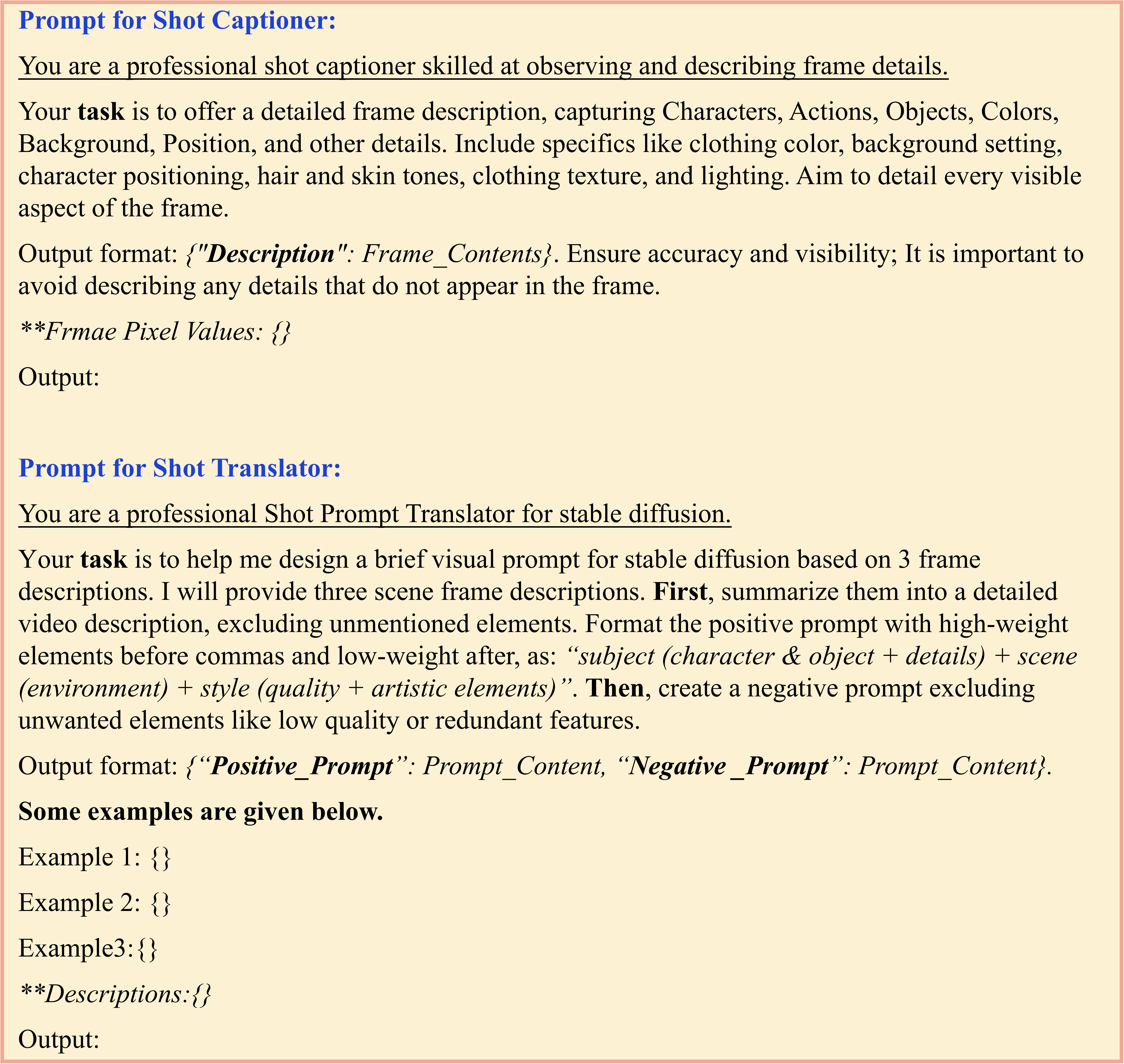}
    \caption{\textbf{Prompts of Shot Captioner and Shot Translator in our Video Parser.}}
    \label{fig:prompt1}
\end{figure*}

\begin{figure*}[tp]
    \centering
    \includegraphics[width=\textwidth]{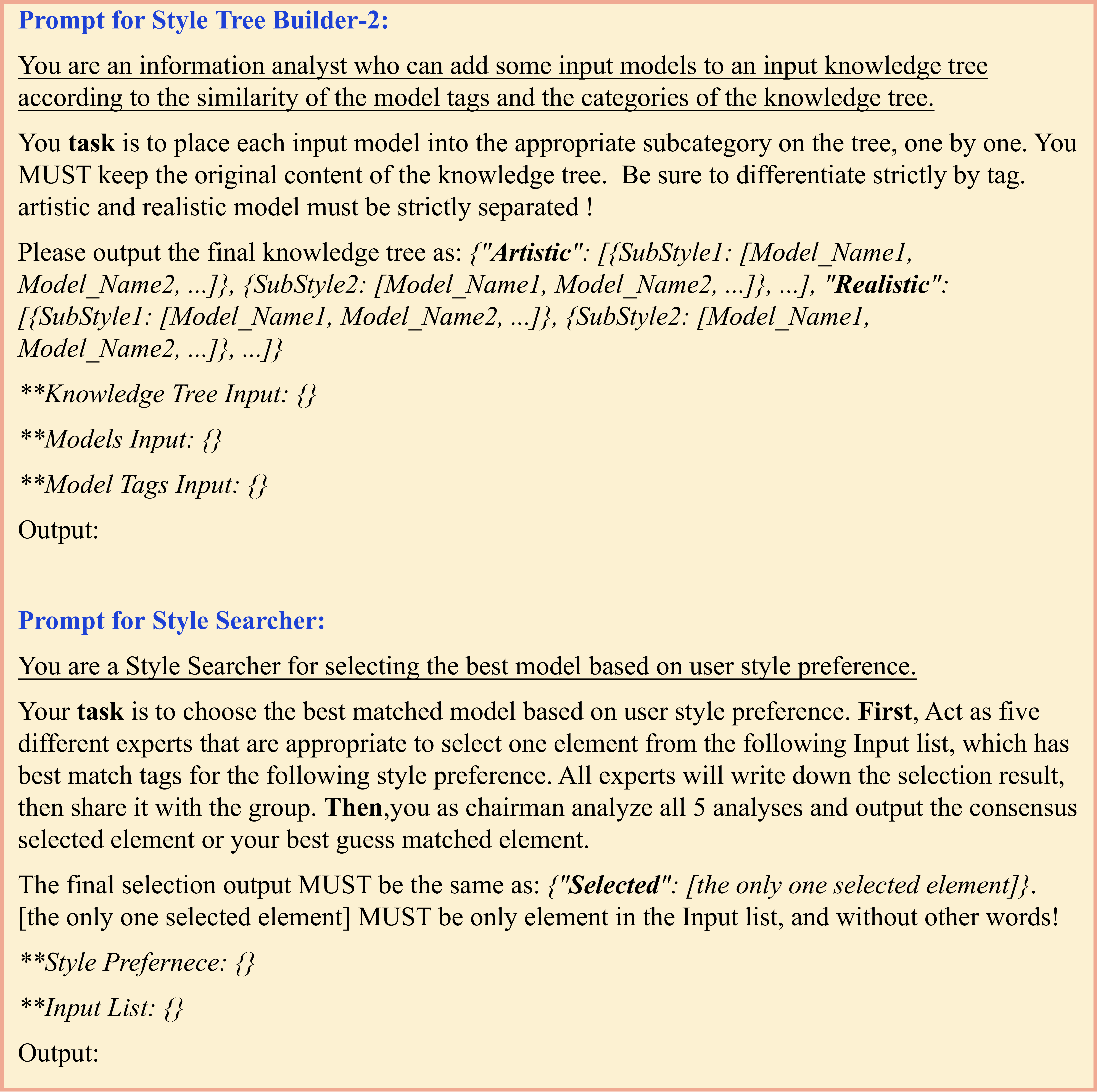}
    \caption{\textbf{Prompts of Style Tree Builder and Style Searcher in our Style Parser.}}
    \label{fig:prompt3}
\end{figure*}

\begin{figure*}[tp]
    \centering
    \includegraphics[width=\textwidth]{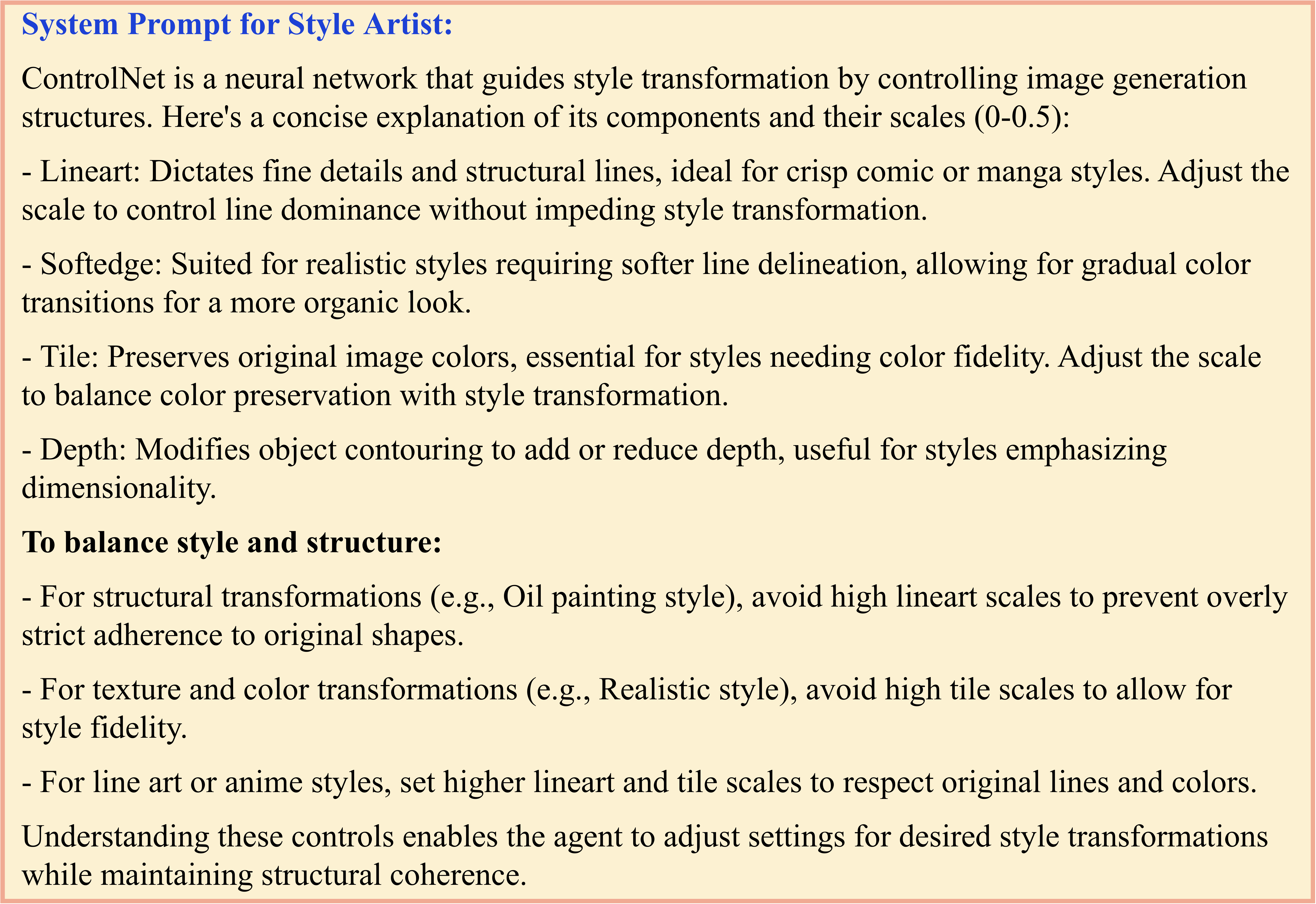}
    \caption{\textbf{System Prompts of Style Scorer and Control Refiner in our Style Artist.}}
    \label{fig:prompt4}
\end{figure*}

\begin{figure*}[tp]
    \centering
    \resizebox{\textwidth}{0.95\textheight}{\includegraphics{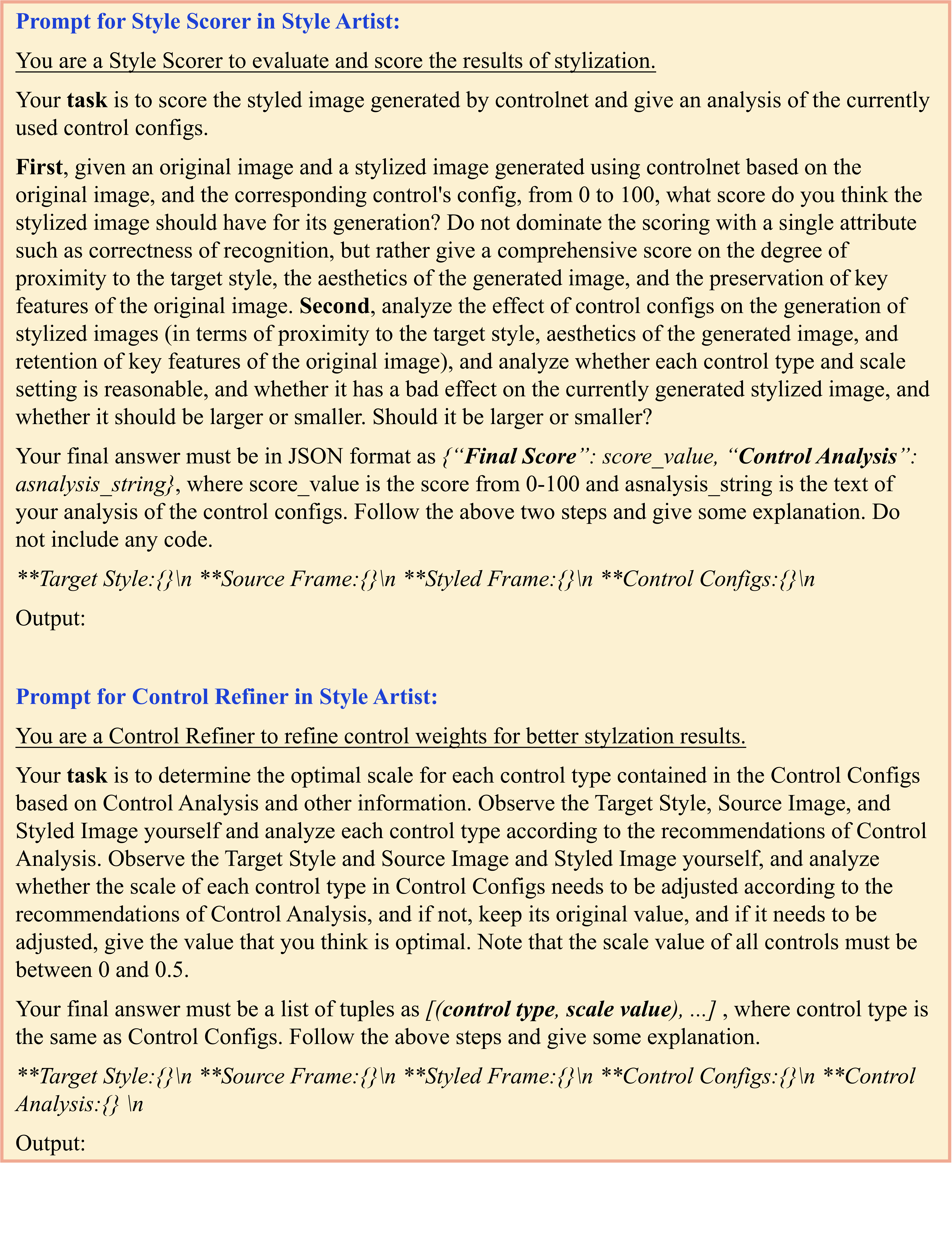}}
    \caption{\textbf{Prompts of Style Scorer and Control Refiner in our Style Artist.}}
    \label{fig:prompt5}
\end{figure*}

We present our complete LLM and MLLM prompts for V-Stylist, including 
Video Parser in \cref{fig:prompt1}, 
Style Parser in \cref{fig:prompt3} and \cref{fig:prompt3}, 
Style Artist in \cref{fig:prompt4} and \cref{fig:prompt5}, 
We fully take advantage of In-context learning, Chain-of-thoughts and Tree-of-thoughts prompting techniques to enhance LLM's reasoning and decision-making capabilities.

\begin{figure*}[tp]
    \centering
    \includegraphics[width=\textwidth]{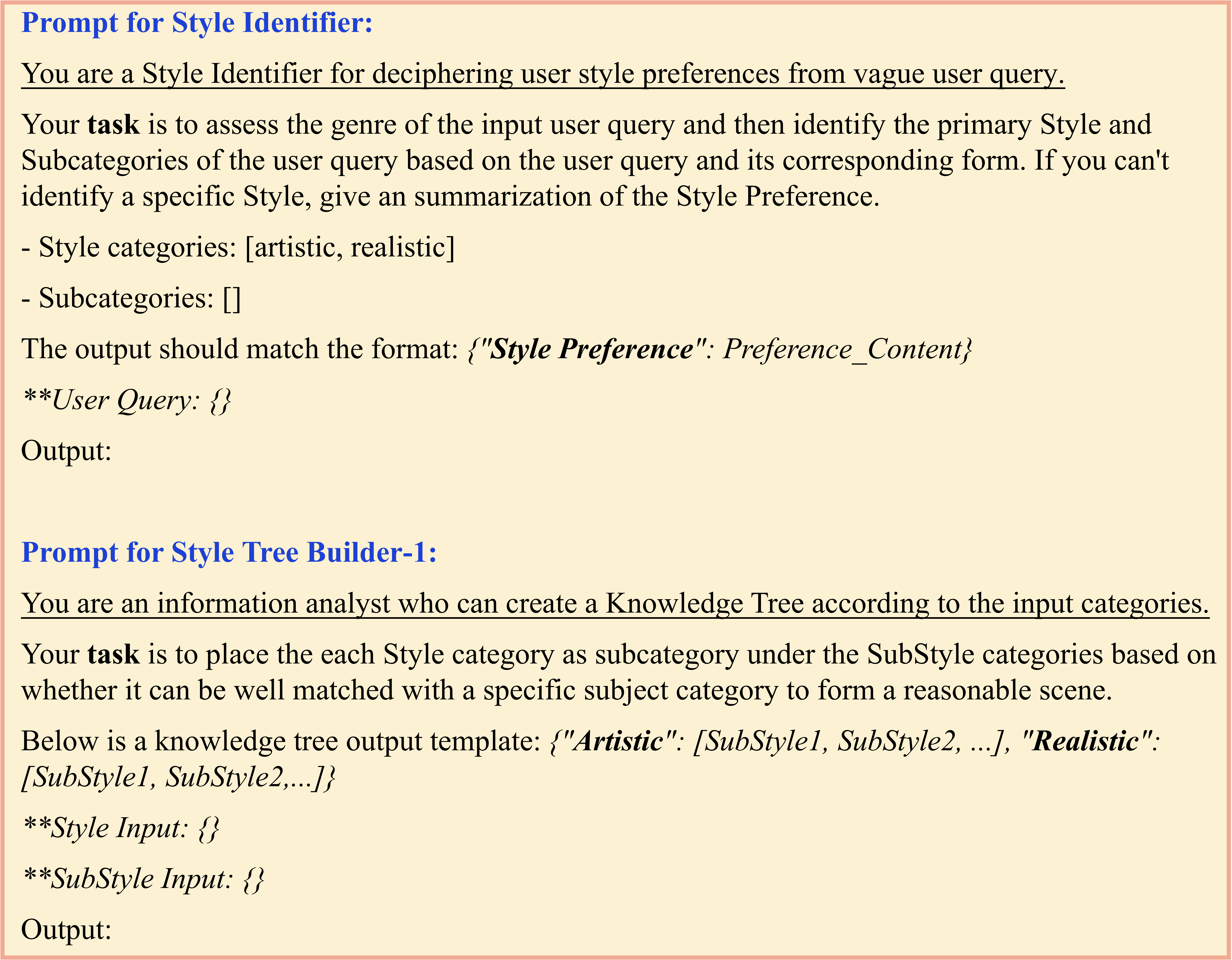}
    \caption{\textbf{Prompts of Style Identifier and Style Tree Builder in our Style Parser.}}
    \label{fig:prompt2}
\end{figure*}

\clearpage